\documentclass[sigconf]{acmart}

\usepackage{booktabs} % 导入 booktabs 宏包
\usepackage{multirow} % 导入 multirow 宏包
\usepackage{colortbl} % For color in tables
\usepackage{enumitem}
\usepackage{xcolor}   % For color definitions
\usepackage{svg}
\usepackage{subcaption}
\usepackage{graphicx}
\AtBeginDocument{%
  }

\acmConference[]{}{}{}
\setcopyright{none}
\renewcommand\footnotetextcopyrightpermission[1]{}
\makeatletter

\begin{document}

\title{RIPCN: A Road Impedance Principal Component Network for Probabilistic Traffic Flow Forecasting}
\thanks{This paper has been accepted for publication at KDD 2026.}

\author{Haochen Lv}
\authornote{Equal contribution.}
\affiliation{%
  \department{School of Computer Science and Technology}
  \institution{Beijing Jiaotong University}
  \city{Beijing}
  \country{China}
}
\email{lvhaochen@bjtu.edu.cn}

\author{Yan Lin}
\authornotemark[1]
\affiliation{%
  \department{Department of Computer Science}
  \institution{Aalborg University}
  \city{Aalborg}
  \country{Denmark}
}
\email{lyan@cs.aau.dk}

\author{Shengnan Guo}
\authornote{Corresponding author.}
\affiliation{%
  \department{School of Computer Science and Technology}
  \institution{Beijing Jiaotong University}
  \city{Beijing}
  \country{China}
}
\affiliation{%
  \institution{Key Laboratory of Big Data \& Artificial Intelligence in Transportation, Ministry of Education}
  \city{Beijing}
  \country{China}
}
\email{guoshn@bjtu.edu.cn}

\author{Xiaowei Mao}
\author{Hong Nie}
\author{Letian Gong}
\affiliation{%
  \department{School of Computer Science and Technology}
  \institution{Beijing Jiaotong University}
  \city{Beijing}
  \country{China}
}
\email{maoxiaowei@bjtu.edu.cn}
\email{hongnie@bjtu.edu.cn}
\email{gonglt@bjtu.edu.cn}

\author{Youfang Lin}
\author{Huaiyu Wan}
\affiliation{%
  \department{School of Computer Science and Technology}
  \institution{Beijing Jiaotong University}
  \city{Beijing}
  \country{China}
}
\affiliation{%
  \institution{Beijing Key Laboratory of Traffic Data Mining and Embodied Intelligence}
  \city{Beijing}
  \country{China}
}
\email{yflin@bjtu.edu.cn}
\email{hywan@bjtu.edu.cn}

\renewcommand{\shortauthors}{Haochen Lv et al.}

\begin{abstract}
Accurate traffic flow forecasting is crucial for intelligent transportation services such as navigation and ride-hailing. In such applications, uncertainty estimation in forecasting is important because it helps evaluate traffic risk levels, assess forecast reliability, and provide timely warnings. As a result, probabilistic traffic flow forecasting (PTFF) has gained significant attention, as it produces both point forecasts and uncertainty estimates. However, existing PTFF approaches still face two key challenges: (1) how to uncover and model the causes of traffic flow uncertainty for reliable forecasting, and (2) how to capture the spatiotemporal correlations of uncertainty for accurate prediction.

To address these challenges, we propose RIPCN, a Road Impedance Principal Component Network that integrates domain-specific transportation theory with spatiotemporal principal component learning for PTFF. RIPCN introduces a dynamic impedance evolution network that captures directional traffic transfer patterns driven by road congestion level and flow variability, revealing the direct causes of uncertainty and enhancing both reliability and interpretability. In addition, a principal component network is designed to forecast the dominant eigenvectors of future flow covariance, enabling the model to capture spatiotemporal uncertainty correlations. This design allows for accurate and efficient uncertainty estimation while also improving point prediction performance.
Experimental results on real-world datasets show that our approach outperforms existing probabilistic forecasting methods. 
\end{abstract}





\maketitle

\section{Introduction}

Traffic flow forecasting is essential for understanding and managing traffic, supporting services like navigation~\cite{dai2020hybrid, shen2025towards}, ride-hailing~\cite{fu2021traffic}, and logistics delivery~\cite{laynes2023framework}. For example, Baidu Maps uses traffic forecasts to help drivers avoid congestion and save time by suggesting better routes or transportation modes~\cite{xia2022dutraffic}.

However, many works focus on point predictions, which fail to provide accurate forecasts when faced with unpredictable situations. In such cases, reliable uncertainty estimation is crucial, as it measures traffic risk levels and model confidence, thus providing early warnings in advance. In this work, we aim to achieve efficient and effective probabilistic traffic flow forecasting, which delivers both point predictions and uncertainty quantification.

Traffic flow uncertainty arises from two main factors. One is related to inherent traffic patterns, such as during peak hours or on key high-traffic road segments, where flow exhibits greater variance, i.e., higher uncertainty~\cite{jin2024spatial}. The other relates to unpredictable risk events, such as traffic accidents, which can suddenly disrupt traffic flow and escalate uncertainty. Both peak hours and traffic accidents lead to significant traffic flow fluctuations, which result in uncertainty. Therefore, the direct cause of uncertainty lies in the underlying \emph{traffic flow fluctuations}.

\begin{figure}[t]
\vspace{-0mm}
  \centering
\includegraphics[width=0.8\linewidth]{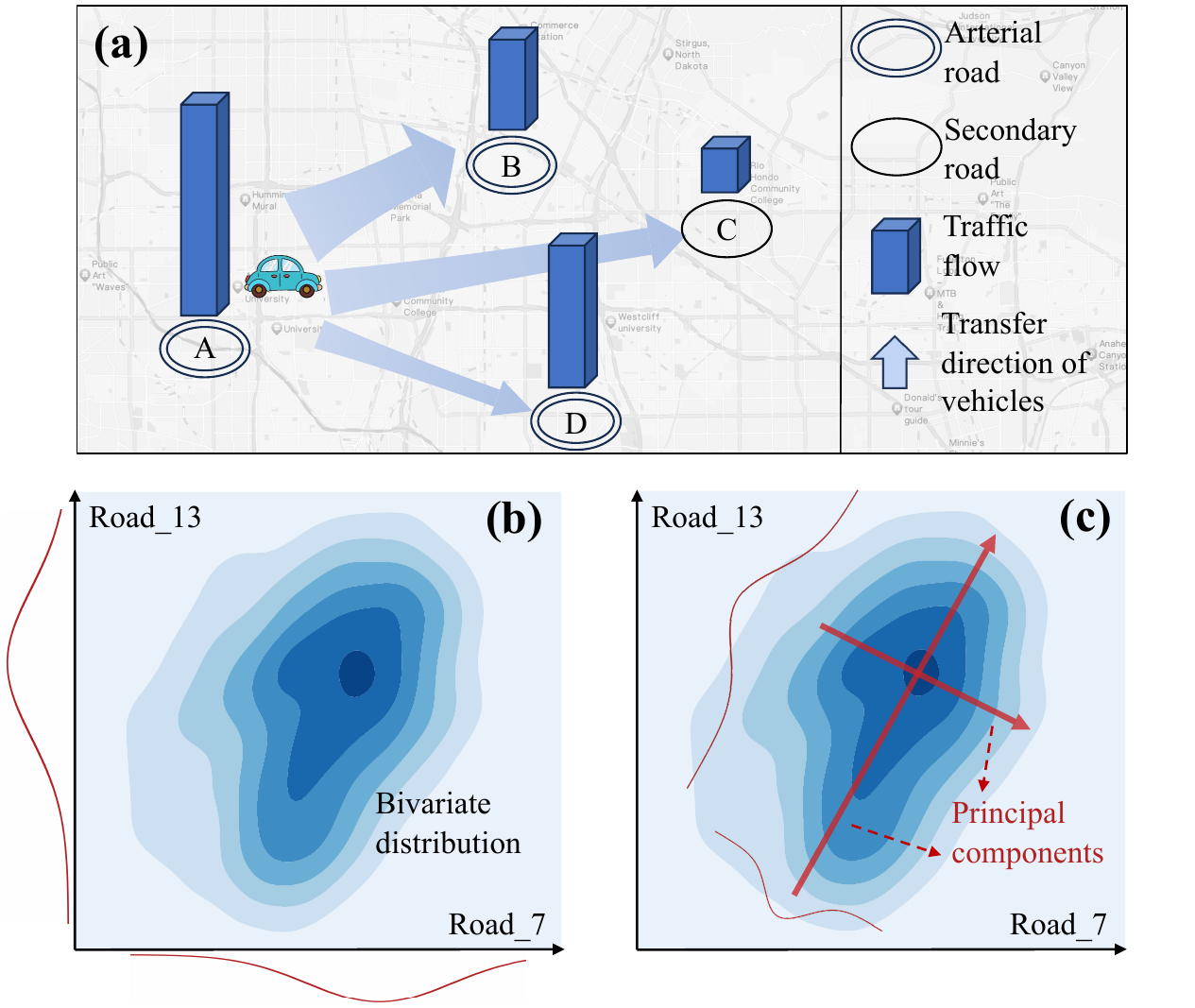}
    \vspace{-0mm}
  \caption{(a) shows the traffic transfer patterns, influenced by traffic flow and road types. (b) shows the distribution of traffic flow in the PEMS08 dataset from 18:00 to 21:00. The red curve represents the uncertainty estimated by independently modeling each road. In contrast, the red curve in (c) illustrates the uncertainty along the principal components.}
  \label{npc-intro}
  \vspace{-0mm}
\end{figure}

To estimate uncertainty, many works focus on \emph{probabilistic traffic flow forecasting}~\cite{salinas2020deepar, rubanova2019latent, wu2021quantifying, an2024spatio}. These methods typically draw samples from probability distributions learned by a model and compute the mean and variance as point and uncertainty estimates~\cite{wu2021quantifying}. Despite this progress, two challenges remain unresolved:

\textbf{First, uncovering and modeling the causes of uncertainty for reliable forecasting poses a significant challenge.}
The direct cause of uncertainty is traffic flow fluctuations, as mentioned above. These fluctuations primarily occur as vehicles move from one road segment to another, transferring their uncertainty across the road network. Therefore, modeling such \emph{traffic transfer patterns} is essential for explaining why and how traffic flow fluctuates. However, existing methods are purely data-driven and lack domain-specific traffic knowledge, making it difficult to model them effectively. For instance, in Fig.~\ref{npc-intro}(a), vehicles in a busy segment \(A\) can choose among three connected segments. Although segment \(C\) currently shows the lowest flow, many vehicles prefer segment \(B\) because it has multiple lanes and higher capacity. This example shows how road capacity shapes traffic transfer patterns and thus affects how uncertainty propagates, which is a principle difficult for purely data-driven methods to capture.

\textbf{Second, effectively and efficiently modeling the spatiotemporal correlations of uncertainty remains a key challenge for accurate forecasting.} Traffic flows across all time steps and road segments can be regarded as a multivariate random variable, and the goal of probabilistic forecasting is to model the joint probability distribution over these variables. The spatiotemporal correlations of uncertainty characterize how uncertainty propagates across both time and space. These correlations are reflected in the shape and orientation of the distribution, as shown in Fig.~\ref{npc-intro}(b), revealing the dominant directions along which uncertainty extends in the joint distribution space.

Most existing methods focus solely on estimating marginal variances, typically defined as $\sigma_X^2 = \mathbb{E}[(X - \mu_X)^2]$, where $X$ denotes the traffic flow on a single road at a specific time step and $\mu_X$ is the expected value. As shown in Fig.~\ref{npc-intro}(b), ignoring the correlations of uncertainty can lead to overestimated variance. To overcome this issue, we aim to model the covariance $\sigma_{XY} = \mathbb{E}[(X-\mu_X)(Y-\mu_Y)]$ between any two traffic flow variables $X$ and $Y$, which quantifies how their fluctuations co-vary across space and time. However, (1) in traffic flow forecasting, the underlying correlations are highly dynamic, making it difficult for a covariance structure derived from historical data to generalize well to future scenarios. (2) Moreover, accurately estimating the full spatiotemporal covariance structure is challenging and computationally expensive due to the high dimensionality of traffic data. 

To address these challenges, we propose the \underline{R}oad \underline{I}mpedance \underline{P}rincipal \underline{C}omponent \underline{N}etwork (\textbf{RIPCN}) for probabilistic traffic flow forecasting. RIPCN consists of two modules: an \emph{impedance evolution network} and a \emph{principal component network}. In the impedance evolution network, we introduce road impedance theory from the field of transportation to model the relationship between traffic flow and traffic transfer behavior by integrating domain-specific factors such as road capacity. By dynamically evolving the impedance over time, we construct a time-varying impedance graph that captures the underlying traffic transfer patterns. Guided by this graph, the principal component network effectively forecasts the eigenvectors of the future covariance structure, referred to as the spatiotemporal principal components, to model the correlations in uncertainty. This approach retains the most critical information in the covariance structure, as illustrated in Fig.~\ref{npc-intro}(c), while significantly reducing computational costs. Moreover, we theoretically and empirically demonstrate that this method can lead to more accurate mean predictions. Our main contributions of this work are as follows:
% \vspace{-1mm}
\begin{itemize}[leftmargin=*]
\item We propose RIPCN, a dual-network architecture that integrates transportation domain knowledge with spatiotemporal principal component learning for probabilistic traffic flow forecasting.
\item RIPCN introduces a dynamic impedance evolution network that captures directional traffic transfer patterns driven by road congestion level and flow variability, revealing the underlying causes of uncertainty and enhancing both reliability and interpretability.
\item  A principal component forecasting network is developed to predict the dominant eigenvectors of future flow covariance, enabling compact and structured modeling of spatiotemporal uncertainty. This design supports accurate and efficient uncertainty quantification, as well as improved point forecasting performance.
\item Theoretical analysis and extensive experiments on real-world traffic datasets demonstrate that RIPCN outperforms existing methods in forecasting accuracy, computational efficiency, and model interpretability.
\end{itemize}

%\vspace{-4mm}
\section{Related Work}
Traffic forecasting has long been a significant area of research. Early studies typically relied on traditional time-series models, such as Historical Average~\cite{hamilton2020time}, ARIMA~\cite{ansley1984estimation}, and VAR~\cite{zivot2006vector}, to model temporal dependencies.

With the rise of deep learning, many works have adopted Convolutional Neural Networks (CNNs) for spatial modeling and Recurrent Neural Networks (RNNs) for temporal dynamics, such as STResNet~\cite{zhang2017deep} and ConvLSTM~\cite{yang2018short}. To better capture the structure of road networks, Graph Neural Networks (GNNs) have been introduced. For instance, DCRNN~\cite{li2018diffusion} combines GRU~\cite{chung2014empirical} with graph convolution, while STGCN~\cite{yu2017spatio} leverages ChebNet~\cite{defferrard2016convolutional}. Attention-based models such as GMAN~\cite{zheng2020gman}, ASTGNN~\cite{guo2021learning}, and PDFormer~\cite{jiang2023pdformer} further enhance spatiotemporal modeling.

Despite the success of these approaches, most traffic flow forecasting models remain deterministic, providing only point estimates. To quantify uncertainty, probabilistic forecasting methods have been proposed that generally assume model parameters or latent representations follow prior probability distributions.

Some methods have estimated the mean and variance via multiple samples. MC Dropout~\cite{wu2021quantifying}, for example, introduces dropout layers and treats each output as a sample from the posterior. DeepAR~\cite{salinas2020deepar} assumes a specific probabilistic form and models the mean and variance jointly. Latent ODE~\cite{rubanova2019latent} leverages ordinary differential equations to capture latent dynamics under a Gaussian assumption. Alternatively, other approaches have directly estimated variance by imposing distributional constraints. For instance, DER~\cite{amini2020deep} and SDER~\cite{wu2024evidence} use evidential priors to infer hyperparameters of the evidential distribution.

More recently, generative models have shown strong potential in probabilistic forecasting due to their ability to model complex distributions. DiffSTG~\cite{zheng2020gman} integrates spatiotemporal GNNs with diffusion models~\cite{ho2020denoising} for traffic forecasting, and Nehme et al.~\cite{nehme2023uncertainty} enhance the efficiency of diffusion processes while explicitly modeling inter-pixel uncertainty in computer vision. STGNF~\cite{an2024spatio} leverages normalizing flows, using invertible mappings to transform complex distributions into tractable ones.

However, most existing methods estimate uncertainty independently across time and space, neglecting the spatiotemporal dependencies in uncertainty that are critical to traffic data, which can lead to overly pessimistic and less informative uncertainty estimates.

% \vspace{-4mm}
\section{Preliminaries}
\label{3}
In this section, we provide the necessary background and theoretical foundation for our proposed approach.

\subsection{Notations and Problem Definition}
\vspace{1mm}
\textbf{Traffic Network.} A traffic network can be modeled as a graph $\mathcal{G} = (\mathcal{V}, \mathcal{E}, \boldsymbol{A})$, where $\mathcal{V}$ is the set of $N$ nodes representing road segments, $\mathcal{E}$ is the set of edges, and $\boldsymbol{A}$ is the adjacency matrix. 

\vspace{2mm}
\noindent \textbf{Probabilistic Traffic Flow Forecasting.} Let $\boldsymbol{X}_t \in \mathbb{R}^{N} $ represent the traffic flow on the road network at time step $t$. Given historical traffic flow data for $\tau$ time steps and road features $\boldsymbol{P}\in \mathbb{R}^{N \times M}$, our goal is to learn a function $\mathcal{F}$ that captures the probability distribution of future traffic flow and outputs the mean $\boldsymbol{\mu}$ and standard deviation $\boldsymbol{\sigma}$ by drawing multiple samples:
%\vspace{-5pt} % 调整公式上方的空白
\begin{equation}
\mathcal{F}: \left( \boldsymbol{X}_{t-\tau+1:t}, \boldsymbol{P}, \mathcal{G} \right) 
\longrightarrow  
\left[ 
\boldsymbol{\mu}\left( \boldsymbol{\hat{X}}_{t+1:t+T} \right),
\ \boldsymbol{\sigma}\left( \boldsymbol{\hat{X}}_{t+1:t+T} \right)
\right],
\label{define}
\end{equation}
where $\boldsymbol{X}_{t-\tau+1:t} \in \mathbb{R}^{\tau \times N}$ denotes the traffic flow observed in the past, which will be simplified as $\boldsymbol{X}_H$ in the following text. Similarly, $\boldsymbol{\hat{X}}_{t+1:t+T}\in \mathbb{R}^{T \times N}$ represents the predicted future traffic flow, simplified as $\boldsymbol{\hat{X}}_P$. 

\subsection{Supporting Theoretical Concepts}
\vspace{1mm}
\textbf{Road Traffic Impedance Theory.} 
\label{impedance theory}
In the field of transportation, the road traffic impedance, simply referred to as road impedance, quantifies the level of congestion on road segments. It is a key factor influencing travelers’ route choices, as they typically prefer routes with lower impedance, which consequently shapes the dynamic distribution of traffic flows across the road network~\cite{he2013discussion}.

To study road impedance, the Bureau of Public Roads derived the BPR function~\cite{qiu2022recalibration, pan2022calibration, gore2023modified}. The BPR function describes road impedance as road travel time: 
\begin{equation}
    T_{a}(X)=t_a\times\left[ 1 + \alpha\left(\frac{X} {C_{a}} \right)^\beta\right],
\label{road-impedance-func}
\end{equation}
where \(T_a(X)\) denotes the travel time on road segment \(a\) and represents the road impedance. The parameter $t_a$ denotes the free-flow travel time, calculated as the road length divided by the speed limit, representing travel under uncongested conditions. For simplicity, roads with similar lengths and speed limits can be assumed to have the same constant $t_a$. The terms \(X\) and \(C_a\) denote the traffic flow and capacity of road segment \(a\), respectively. The parameters $\alpha$ and $\beta$ are typically non-negative constants set to 0.15 and 4.

\vspace{0mm}
\noindent \textbf{Principal Component Analysis.} 
Principal Component Analysis (PCA) is a widely used technique for data dimensionality reduction. The core idea of PCA is to identify a new coordinate system where the first few dimensions, known as principal components (PCs), capture most of the variance in the original dataset. The mathematical formulation of the PCs is given as follows: 
\begin{equation}
\begin{aligned}
\boldsymbol{w}_k &= \underset{\boldsymbol{w}}{\arg\max} \; \frac{1}{n} \sum_{i=1}^{n} \left( \boldsymbol{x}_i^\top \boldsymbol{w} \right)^2 \\
& \text{subject to} \quad \|\boldsymbol{w}\|_2 = 1, \quad \boldsymbol{w}^\top \boldsymbol{w}_j = 0 \quad \text{for } j = 1, \dots, k-1
\end{aligned}
\label{tnpc}
\end{equation}
where $\boldsymbol{x}_i$ denotes the $i^{\text{th}}$ centered sample data, obtained by subtracting the mean from the original data. The quantity $\frac{1}{n} \sum_{i=1}^{n} (\boldsymbol{x}_i^\top \boldsymbol{w})^2$ represents the variance of the data along the direction $\boldsymbol{w}$.

\begin{figure*}[h]
  \centering
  \includegraphics[width=\linewidth]{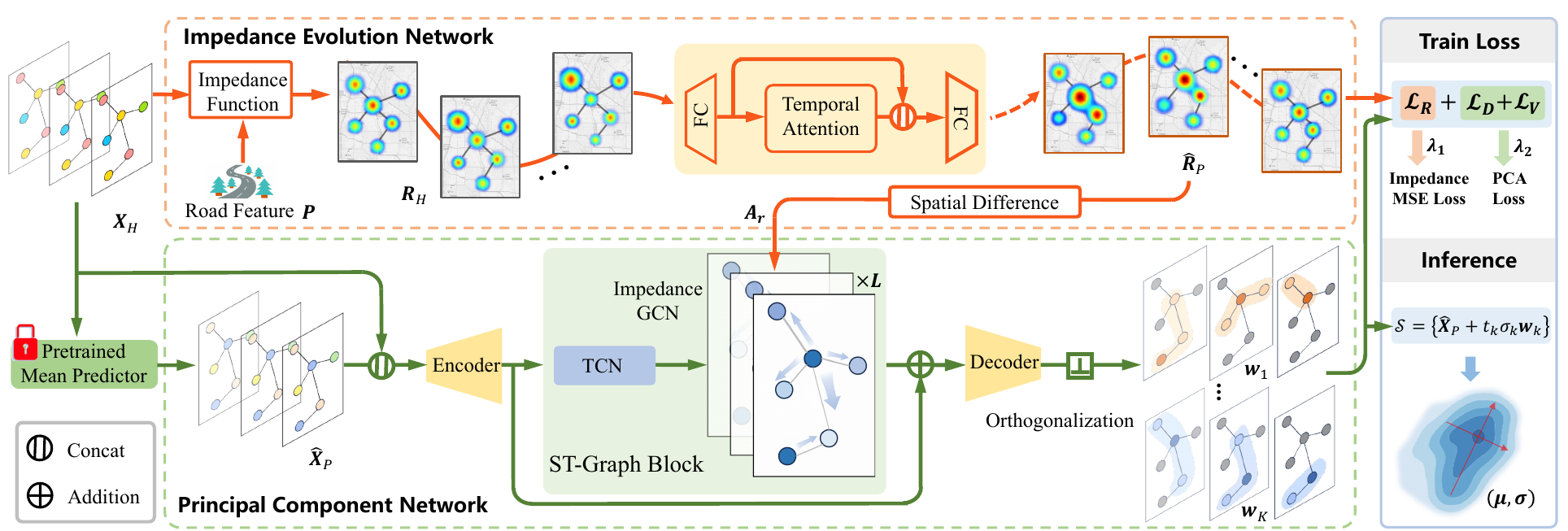}
  \vspace{-5mm}
  \caption{The overall architecture of RIPCN. The impedance evolution network takes historical flow to evolve road impedance, generating a dynamic impedance graph. The principal component network leverages this graph to predict the PCs.}
  \Description{overall architecture}
  \vspace{-0mm}
  \label{architecture}
\end{figure*}
%%%%%%%%%%%%%%%%%%%%%%%%%%%%%%%%%%%%%%%%%%%%%%%

\section{The RIPCN Model}

\subsection{Overview}

In this study, we propose RIPCN, a novel architecture for probabilistic traffic flow forecasting, as illustrated in Fig.~\ref{architecture}. RIPCN comprises two networks: the impedance evolution network and the principal component network. The impedance evolution network evolves historical road impedance $\boldsymbol{R}_H$ into future road impedance $\boldsymbol{R}_P$ through a temporal attention block, and subsequently generates a dynamic impedance graph $\boldsymbol{A}_r$. This graph is then input into the principal component network to guide the model in understanding traffic transfer patterns. The principal component network predicts spatiotemporal principal components (PCs) by processing input flows through $L$ spatiotemporal graph blocks (ST-Graph Blocks), which combine TCN and GCN with a dynamic impedance graph, followed by orthogonalization. RIPCN is trained by impedance MSE loss and PCA loss. In inference, by combining the PCs with $\boldsymbol{\hat{X}}_P$, a series of samples is generated for probabilistic forecasting.

The following sections introduce the impedance evolution and principal component modules, each with its theoretical framework and corresponding network implementation.

\subsection{Impedance Evolution Module}
\label{4.2}
To uncover and model the causes of traffic flow uncertainty, we incorporate domain-specific knowledge from road impedance theory. By capturing how traffic propagates across road segments, this theory characterizes flow fluctuations, which are the direct source of uncertainty. Such guidance enables the model to learn transfer patterns driven by road congestion and flow variability, thereby improving uncertainty estimation and enhancing interpretability.
\subsubsection{Road Impedance with Traffic Flow Uncertainty}
The BPR function (Eq.~\ref{road-impedance-func}) estimates road impedance based on road capacity and free-flow travel time. However, it does not consider that the intensity of traffic flow fluctuations can also significantly influence actual travel time, in addition to the absolute flow volume. In practice, road segments with higher variability in traffic flow tend to experience greater uncertainty in travel time. To address this, we introduce a flow-based uncertainty factor into the impedance function:
\begin{equation}
T_{a}(X) = t_a \times \left[ 1 + \alpha \left( \frac{X}{C_a} \right)^\beta \right] \times \left(1 + \frac{\sigma_a}{\mu_a} \right),
\label{impedance_func_new}
\end{equation}
where \(T_a(X)\) denotes the travel time on road segment \(a\). The term \(\sigma_a / \mu_a\) is the coefficient of variation of the traffic flow over the historical $\tau$ time steps, capturing the relative fluctuations. These features \(t_a\), \(C_a\), \(\mu_a\) and \(\sigma_a\) are treated as static or slowly varying attributes associated with each road segment, corresponding to \(\boldsymbol{P} \in \mathbb{R}^{N \times 4}\) as defined in Eq.~\ref{define}.

Importantly, obtaining road capacity $C_a$ is challenging because detailed information, ideally sourced from road construction manuals, is often unavailable. Transportation studies such as~\cite{minderhoud1997assessment, brilon2007implementing} typically estimate $C_a$ using complex models based on headways, traffic counts, and other factors. To ensure our method remains applicable in data-scarce scenarios, we propose a simplified estimation of road capacity:
% using historical speed and occupancy statistics, as follows
\begin{equation}
    C_a = X_{\mathrm{max}} \times \left[ 
        1 + \frac{1}{2} \left( 
            \frac{s_a}{s_{\mathrm{mean}}} + \frac{o_{\mathrm{mean}}}{o_a} 
        \right) 
    \right],
    \label{Ca}
\end{equation}
where $X_\mathrm{max}$ is the maximum observed traffic flow on road segment $a$, $s_a$ and $o_a$ represent the vehicle speed and occupancy at the time when $X_\mathrm{max}$ is recorded, and $s_\mathrm{mean}$ and $o_\mathrm{mean}$ denote the average speed and occupancy across all road segments. 

This formulation suggests that the maximum observed flow can serve as a baseline estimate of capacity. If the corresponding speed $s_a$ is above the average or the occupancy $o_a$ is below the average, it implies that the segment may support a higher capacity than $X_\mathrm{max}$. Conversely, lower speed or higher occupancy indicates the road is operating near its capacity limit.

Furthermore, in cases where only historical traffic flow data is available, $X_\mathrm{max}$ alone can serve as a proxy for $C_a$. We empirically validate the effectiveness of these approximation strategies.

\subsubsection{Impedance Evolution Network}
Although the impedance function Eq.~\ref{impedance_func_new} incorporates a variation factor, it still remains largely static. Therefore, to guide the prediction of future traffic dynamics, we propose an Impedance Evolution Network that models the temporal progression of road impedance using historical observations. 

Specifically, historical road impedance is first computed using Eq.~\ref{impedance_func_new}, and subsequently passed through a fully connected encoder that transforms the raw sequence into a compact representation suited for modeling its temporal evolution.
\begin{equation}
\boldsymbol{R}_H = \text{ImpedanceFunc}(\boldsymbol{X}_H, \boldsymbol{P}),
\quad
\boldsymbol{R}_H^h = \text{Encoder}(\boldsymbol{R}_H),
\end{equation}
where \( \boldsymbol{X}_H \in \mathbb{R}^{\tau \times N \times 1} \) denotes the historical traffic flow data, \( \boldsymbol{R}_H \in \mathbb{R}^{\tau \times N \times 1} \) represents the corresponding computed road impedance, and \( \boldsymbol{R}_H^h \in \mathbb{R}^{\tau \times N \times F} \) is the encoded representation.

Next, we employ temporal attention~\cite{vaswani2017attention} to infer future road impedance based on historical values. The attention weights across time steps are computed as follows:
\begin{equation}
\resizebox{.90\linewidth}{!}{
\begin{math}
\begin{split}
    \alpha_{i,j}  = \frac{ \exp\left( \left( \boldsymbol{\mathit{W}}^Q \boldsymbol{\mathit{R}}_H^i \right)^{\top} \left( \boldsymbol{\mathit{W}}^K \boldsymbol{\mathit{R}}_H^j \right) \right)}{\sum_{t=1}^{T} \exp\left( \left( \boldsymbol{\mathit{W}}^Q \boldsymbol{\mathit{R}}_H^i \right)^{\top} \left( \boldsymbol{\mathit{W}}^K \boldsymbol{\mathit{R}}_H^j \right) \right)},~~
    \boldsymbol{\mathit{R}}_P^i  = \sum_{j=1}^{T} \alpha_{i,j} \left( \boldsymbol{\mathit{W}}^V \boldsymbol{\mathit{R}}_H^i \right),
\end{split}
\end{math}}
\end{equation}
where $\boldsymbol{W}^{Q},\boldsymbol{W}^{K},\boldsymbol{W}^{V} \in \mathbb{R}^{F\times F}$ are learnable parameters, and $\boldsymbol{R}_P^i$ represents the impedance at the $i$-th future time step. The road impedance at all time steps can be expressed as $\boldsymbol{R} = [\boldsymbol{R}_H, \boldsymbol{R}_P]$.

According to road impedance theory, higher impedance indicates heavier congestion, making it harder for vehicles to enter a segment and easier for traffic to exit. Consequently, the impedance difference between segments captures a directional tendency in traffic flow and acts as a driving force for its redistribution~\cite{wardrop1952road, wang2023abnormal}. This property naturally informs the construction of the road graph, enabling traffic flow transfer patterns to be effectively captured. Therefore, based on actual network connectivity, we compute the impedance difference between each road segment and its neighbors at each time step to generate a dynamic impedance graph that reflects the temporal evolution of directional flow tendencies:
\begin{equation}
    \boldsymbol{\hat{A}}_r^{a,b}= -\left[\text{FC}(\boldsymbol{R}_a)-\text{FC}(\boldsymbol{R}_b)\right] \times \mathbb{I}_{\boldsymbol{{A}}^{a,b}}\text{ },
\end{equation}
where $\text{FC(·)}$ is a fully connected network that decodes the impedance representation $\boldsymbol{R}_a\in \mathbb{R}^{(\tau+T) \times F}$ and $\boldsymbol{R}_b\in \mathbb{R}^{(\tau+T)\times F}$ into impedance values,  ${\boldsymbol{A}}^{a,b}$ represents the connectivity between road segments $a$ and $b$, and $\mathbb{I}_{\boldsymbol{A}^{a,b}}$ is an indicator that returns $1$ if $\boldsymbol{A}^{a,b}$ is positive, $0$ otherwise. 

During training, we use a mean squared error (MSE) loss to encourage the predicted impedance $\boldsymbol{\hat{R}}_P$ to approximate the ground truth impedance \( \boldsymbol{R}_P \), which is derived from the actual traffic flow:
\begin{equation}
    \mathcal{L}_R = \left\| \boldsymbol{R}_P - \boldsymbol{\hat{R}}_P \right\|_2^2
\end{equation}

In summary, the impedance evolution network learns to generate a dynamic impedance graph \( \boldsymbol{\hat{A}}_r \in \mathbb{R}^{(\tau+T) \times N \times N} \), which is subsequently integrated into the principal component network to inform the modeling of spatiotemporal traffic transfer patterns.

\subsection{Principal Component Module}
As shown in Fig.~\ref{npc-intro}, ignoring spatiotemporal correlations and modeling uncertainties independently often leads to overly diffuse estimates. To address this, we adopt a PCA-based approach that predicts the principal components of future uncertainty, providing a compact representation of the covariance structure. This low-rank approximation captures the dominant spatiotemporal variations, reduces the cost and noise of direct covariance estimation, and improves both uncertainty quantification and point forecast accuracy.

\subsubsection{Spatiotemporal Principal Component Forecasting}
\label{4.3.1}
To capture the spatiotemporal correlations of traffic flow uncertainty, covariance is introduced. Covariance is a fundamental statistical measure that quantifies the joint variability between two random variables, characterizing the extent to which they change together. 

For the target traffic flow $\boldsymbol{X}_P \in \mathbb{R}^{T \times N}$, we define the centered data as $ \boldsymbol{X} = \boldsymbol{X}_P - \boldsymbol{\hat{X}}_P $, where $\boldsymbol{\hat{X}}_P$ is the mean predicted by a network. This centering process removes bias and highlights the underlying fluctuations. The covariance structure is then represented by a second-order tensor $\boldsymbol{\Sigma} =\boldsymbol{X} \otimes \boldsymbol{X} \in \mathbb{R}^{T \times N \times T \times N}$, where \(\otimes\) denotes the outer product over both temporal and spatial dimensions. This formulation captures the spatiotemporal dependencies in traffic flow fluctuations.

However, (1) in traffic flow forecasting, the underlying correlations are highly dynamic, making it difficult for a covariance structure derived from historical data to generalize well to future scenarios. (2) Moreover, accurately estimating the full spatiotemporal covariance tensor is both challenging and computationally expensive due to the high dimensionality of traffic data.

\begin{figure}[t]
\vspace{-0mm}
  \centering
\includegraphics[width=1\linewidth]{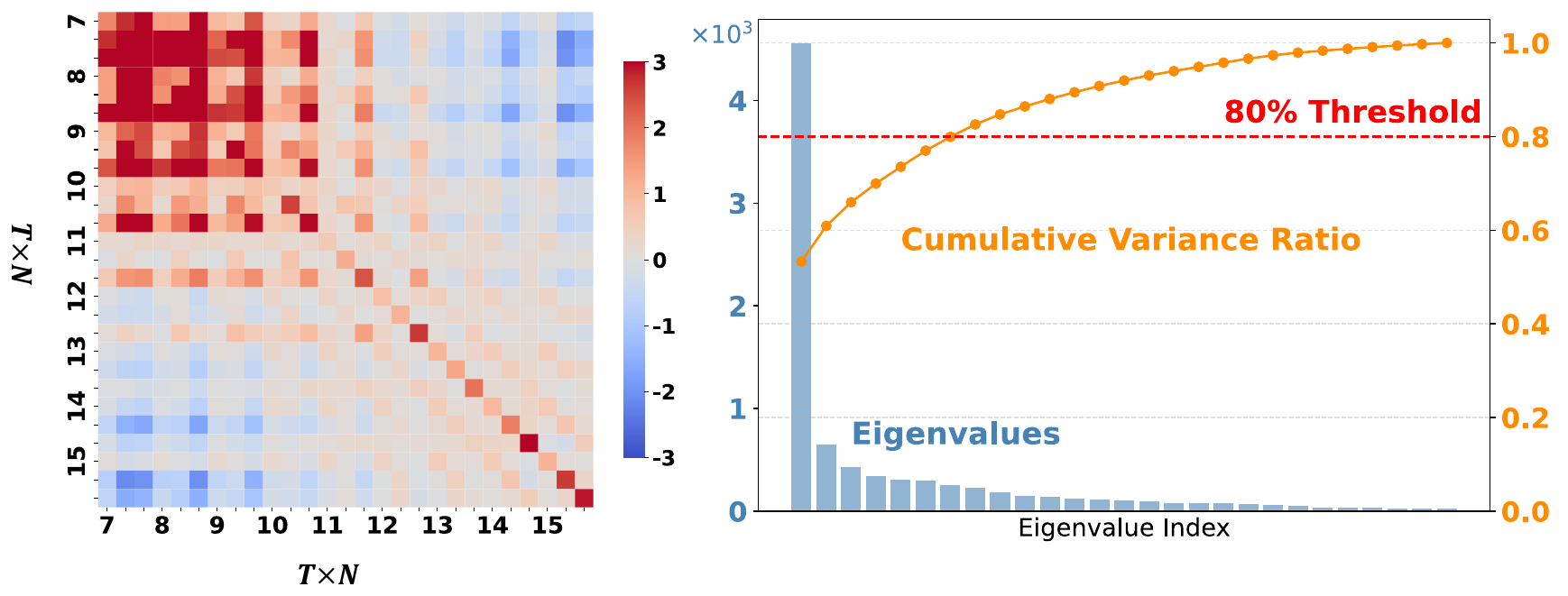}
  \vspace{-0mm}
  \caption{The left figure shows the covariance structure among adjacent segments 7, 13, and 41 in the PEMS08 dataset from 7:00 to 16:00. The right figure shows the corresponding eigenvalue spectrum, illustrating that most of the variance is captured by a few principal components.}
  \label{covariance}
  \vspace{-0mm}
\end{figure}

As shown in Fig.~\ref{covariance}, the covariance of traffic flow varies over time and grows quadratically with the prediction horizon $T$ and the number of segments $N$. Fortunately, we observe that after eigenvalue decomposition, the leading eigenvectors capture the majority of the covariance information.

Based on the above observation, we propose to predict the spatiotemporal eigenvectors of the covariance tensor, referred to as the spatiotemporal principal components (PCs), which can be leveraged to efficiently approximate the underlying covariance structure while effectively suppressing noise and other irrelevant variations:
\begin{equation}
\boldsymbol{\Sigma} \approx \sum_{k=1}^{K} \lambda_k \, \boldsymbol{w}_k \otimes \boldsymbol{w}_k,
\label{cov-eig}
\end{equation}
where $\boldsymbol{w}_k \in \mathbb{R}^{T \times N}$ denotes the $k$-th principal component, $\lambda_k$ is the corresponding eigenvalue, and $K$ is the number of retained components.

Unlike traditional PCA which reduces the dimension of the observed data, our aim is to predict the PCs of future traffic flow due to the dynamic nature of traffic data. Therefore, we parameterize these spatiotemporal PCs by a Principal Component Network, detailed in Sec. \ref{PC network}. The $k$-th output of the network is denoted as $\boldsymbol{d}_k(\boldsymbol{X}_H, \boldsymbol{\hat{X}}_P) \in \mathbb{R}^{T \times N}$. To ensure that the predicted components satisfy the orthogonality and unit norm properties of PCs, we apply the Schmidt orthogonalization procedure. The resulting orthogonal output is denoted as $\boldsymbol{w}_k$, given by:
\begin{equation}
    \tilde{\boldsymbol{w}}_k = \boldsymbol{d}_k - \sum_{i=1}^{k-1} \langle \boldsymbol{d}_k, \boldsymbol{w}_i \rangle_F \cdot \boldsymbol{w}_i, \quad 
    \boldsymbol{w}_k = \frac{\tilde{\boldsymbol{w}}_k}{\|\tilde{\boldsymbol{w}}_k\|_F},
\label{eq:schmidt}
\end{equation}
where $\langle \cdot, \cdot \rangle_F$ denotes the Frobenius inner product, and $\|\cdot\|_F$ is the corresponding Frobenius norm.

Based on the property of PCs shown in Eq.~\ref{tnpc}, we introduce a \textbf{Directional Constraint Loss} to encourage $\boldsymbol{w}_k$ to align with the direction of greatest variation in the traffic flow: 
\begin{equation}
\mathcal{L}_{D} = - \sum_{k=1}^{K} \left\langle \boldsymbol{w}_k, \boldsymbol{X} \right\rangle_F^2,
\label{lw}
\end{equation}
where $\left\langle \boldsymbol{w}_k, \boldsymbol{X} \right\rangle_F^2$ represents the variance of $X$ in the direction of $\boldsymbol{w}_k$. 
Building upon this formulation, we further provide a theoretical justification for its effectiveness by revealing its connection to the covariance structure. To enforce the unit norm constraint \( \| \boldsymbol{w}_k \|_F^2 = 1 \), we introduce the Lagrange multiplier \( \theta_k \) and construct the following Lagrangian:
\begin{equation}
\mathcal{L}(\boldsymbol{w}_k, \theta_k) = - \langle \boldsymbol{w}_k, \boldsymbol{X} \rangle_F^2 + \theta_k (\| \boldsymbol{w}_k \|_F^2 - 1)
\end{equation}
Taking the derivative of \( \mathcal{L} \) with respect to \( \boldsymbol{w}_k \), we obtain:
\begin{equation}
\frac{\partial \mathcal{L}}{\partial \boldsymbol{w}_k} = - 2 \langle \boldsymbol{w}_k, \boldsymbol{X} \rangle_F \cdot \boldsymbol{X} + 2\theta_k \boldsymbol{w}_k
\end{equation}
Setting the gradient to zero, we get the following:
\begin{equation}
\langle \boldsymbol{w}_k, \boldsymbol{X} \rangle_F \cdot \boldsymbol{X} =(\boldsymbol{X} \otimes \boldsymbol{X}) \cdot \boldsymbol{w}_k = \boldsymbol{\Sigma} \cdot \boldsymbol{w}_k = \theta_k \boldsymbol{w}_k
\end{equation}
This confirms that the theoretically optimal solution \( \boldsymbol{w}_k \) is a spatiotemporal eigenvector of the covariance structure \( \boldsymbol{\Sigma} \), satisfying Eq.~\ref{cov-eig}.

While PCs capture the directions of traffic flow variation, we also need to quantify the level of fluctuations along these directions, which is reflected in the variance of the data on each PC. To achieve this, we employ a \textbf{Variance Magnitude Loss} that constrains the norm of orthogonalized outputs $\tilde{\boldsymbol{w}}_k$ to match the variance $\langle \boldsymbol{w}_k, \boldsymbol{X} \rangle_F^2$, via a mean squared error:
\begin{equation}
        \mathcal{L}_{V} =  \sum_{k=1}^{K} \left( \langle \boldsymbol{w}_k, \boldsymbol{X} \rangle_F^2 - \left\| \tilde{\boldsymbol{w}}_k \right\|_F^2 \right)^2
\label{lc}
\end{equation}
Thus, the predicted variance $\sigma_k^2$ at test time is given by $\left\|\tilde{\boldsymbol{w}}_k \right\|_F^2$. Note that this reflects variance along the PC, not the overall variance of the data.

Since the PCs capture the dominant directions of joint spatiotemporal variation in the data, they can be used to model the predictive uncertainty. To this end, we construct a set of samples by extending the predicted mean \( \boldsymbol{\hat{X}}_P \) along each principal direction:
\begin{equation}
    \mathcal{S} = \left\{ \boldsymbol{\hat{X}}_P + t_k \sigma_k \boldsymbol{w}_k \,\middle|\, k = 1, \dots, K \right\},
\label{sample-process}
\end{equation}
where \( \mathcal{S} \) denotes the sample set, and \( t_k \) is a scalar coefficient that adjusts both the magnitude and direction along \( \boldsymbol{w}_k \). This accounts for the fact that each \( \boldsymbol{w}_k \) may either align with or oppose the true direction of variation. We determine the value of \( t_k \) using a binary search on the validation set. We compute the mean and variance of the samples in \( \mathcal{S} \) to obtain the final point prediction $\boldsymbol{\mu}$ and the associated uncertainty estimate $\boldsymbol{\sigma}$. 

Notably, the mean derived from \( \mathcal{S} \) is often more accurate than \( \boldsymbol{\hat{X}}_P \). This improvement is attributed to the favorable properties of the spatiotemporal principal components and our sampling mechanism. We provide both theoretical justification and empirical evidence for this observation in Sec.~\ref{case} and Apps.~\ref{correction theory}.

\subsubsection{Principal Component Network}
\label{PC network}
In the previous section, we focus on the loss constraints necessary for predicting spatiotemporal principal components. Beyond the loss design, a suitable network architecture is also crucial. To this end, we construct a dedicated Principal Component Network tailored for traffic flow data. 

In this network, we first deploy a pretrained mean predictor that forecasts future traffic flow based on historical data. The predicted mean is then concatenated with the historical traffic flow data and input into an encoder, which is implemented by a fully connected layer to extract hidden representations:
\begin{equation}
\begin{split}
\boldsymbol{\hat{X}}_P & = \text{MeanPredictor}(\boldsymbol{X}_H, \mathcal{G}), \\
\boldsymbol{h}_C & = \text{Encoder}(\boldsymbol{X}_H \,||\, \boldsymbol{\hat{X}}_P),
\end{split}
\end{equation}
where $\boldsymbol{\hat{X}}_P \in \mathbb{R}^{T\times N}$ represents the mean of predicted future traffic data, and $\boldsymbol{h}_C\in\mathbb{R}^{T\times N\times F}$ is the hidden representation of the concatenated data. Then $L$ spatiotemporal graph blocks (ST Graph Blocks) are implemented to learn traffic transfer patterns under the guidance of road impedance. Each block leverages a Temporal Convolutional Network (TCN)~\cite{bai2018empirical} to model temporal dependencies. The TCN is implemented as a 1D causal convolution. This operation can be represented as:
\begin{equation}
    \boldsymbol{h}_{\mathcal{T}}=\text{TCN}(\boldsymbol{W}^T,\boldsymbol{h}_C),
\end{equation}
where $\boldsymbol{W}^{T}\in \mathbb{R}^{k\times F\times F^{\prime}}$ represents convolution kernel with kernel size $k$, and $\boldsymbol{h}_{\mathcal{T}}\in\mathbb{R}^{T\times N\times F^{\prime}}$ is the temporal feature representation. The spatial dependencies are captured by Graph Convolutional Network (GCN)~\cite{zhou2020graph} based on dynamic impedance graph $\boldsymbol{A}_r$:
\begin{equation}
   \boldsymbol{h}_{\mathcal{G}}=\sigma((\overline{\boldsymbol{A}}||\boldsymbol{A}_r)\boldsymbol{h}_{\mathcal{T}}\boldsymbol{W}^G),
\end{equation}
where $\boldsymbol{\overline{A}}$ is the normalized adjacency matrix of the road network, $\boldsymbol{W}^G\in\mathbb{R}^{F^{\prime} \times F^{\prime}}$ is learnable parameters and $\boldsymbol{h}_{\mathcal{G}}\in\mathbb{R}^{T\times N\times F^{\prime}}$. In addition to the direct connections between the ST Graph Blocks, skip connections are employed. The decoder in the Principal Component Network is implemented with TCN and fully connected layers to output $\boldsymbol{d}\in \mathbb{R}^{K\times T\times N}$ where $K$ is the number of principal components to be predicted. Finally, after the Schmidt orthogonalization procedure, the PCs are obtained as $\boldsymbol{w}\in \mathbb{R}^{K\times T\times N}$:
\begin{equation}
\boldsymbol{d} = \text{Decoder}(\boldsymbol{h}_{\mathcal{G}}), 
\quad
\boldsymbol{w} = \text{Orthogonalization}(\boldsymbol{d})
\end{equation}

\subsection{Model Training}
To train the RIPCN network, we define the overall loss function as:
\begin{equation}
    \mathcal{L} = \lambda_1 \mathcal{L}_R + \lambda_2 \left( \mathcal{L}_D + \mathcal{L}_V \right),
    \label{overall loss}
\end{equation}
where $\mathcal{L}_R$ supervises the dynamic impedance, while $\mathcal{L}_D$ and $\mathcal{L}_V$ together constitute the \textbf{PCA loss}, guiding the model to learn the principal components and the variance along those components. The coefficients $\lambda_1$ and $\lambda_2$ are hyperparameters. During training, we first fix $\lambda_1 = 1$ to prioritize learning the road impedance. Then, $\lambda_2$ is gradually increased from 0 to 1 in a linear schedule, encouraging the model to capture the principal components and their variances under the guidance of impedance learning.

\section{Experiments}
\subsection{Data Description}
We conduct experiments on four real-world traffic datasets: PEMS03, PEMS04, PEMS08~\cite{chen2001freeway}, and Seattle~\cite{cui2018deep, guo2023self}. The PEMS03 dataset contains only traffic flow characteristics, while the other three datasets include traffic speed, occupancy, and flow characteristics. Detailed statistics of these datasets are in Tab.~\ref{data-description}.
\setlength{\tabcolsep}{6pt}
\begin{table}[h]
\centering
\vspace{-0mm}
\caption{Dataset Description.}
\vspace{0mm}
\label{data-description}
\begin{tabular}{ccccc}
\toprule
 \multicolumn{1}{c}{\bf Properties}  &\multicolumn{1}{c}{\bf PEMS03} &\multicolumn{1}{c}{\bf PEMS04}  &\multicolumn{1}{c}{\bf PEMS08}  &\multicolumn{1}{c}{\bf Seattle} \\
\midrule
Time range       & 3 months & 2 months & 2 months & 1 year \\
Time interval    & 5 min & 5 min       & 5 min       & 1 h         \\
\# of nodes      & 358 & 307         & 170         & 323         \\
Mean of flow     & 179  & 207       & 229       & 879     \\
Std of flow    & 143   & 156      & 145      & 525    \\
\bottomrule
\end{tabular}
\vspace{0mm}
\end{table}

% \begin{table}[h]
% \centering
% \caption{Dataset Description.}
% \vspace{-3mm}
% \label{data-description}
% \begin{tabular}{cccccc}
% \toprule
% \bf Dataset & \bf Time range & \bf Time interval & \# of nodes & \bf Mean & \bf Std \\
% \midrule
% \bf PEMS03     & Sep - Nov 2018 & 5 min            & 358         & 179 veh/5min & 143 veh/5min \\
% \bf PEMS04     & Jan - Feb 2018 & 5 min            & 307         & 207 veh/5min & 156 veh/5min \\
% \bf PEMS08     & Jul - Aug 2016 & 5 min            & 170         & 229 veh/5min & 145 veh/5min \\
% \bf Seattle    & Jan - Dec 2015 & 1 h              & 323         & 879 veh/h    & 525 veh/h    \\
% \bottomrule
% \end{tabular}
% \vspace{-5mm}
% \end{table}
\subsection{Experimental Settings}
\textbf{Baselines and Metrics.}
We compare our RIPCN with nine advanced probabilistic forecasting methods, which can be categorized into three groups based on their uncertainty modeling paradigms. Approximate Bayesian inference methods include MC Dropout~\cite{wu2021quantifying}, DeepAR~\cite{salinas2020deepar}, and Latent ODE~\cite{rubanova2019latent}, which introduce stochasticity into models to approximate the posterior via multiple sampling. Generative probabilistic models, including STGNF~\cite{an2024spatio}, PriSTI~\cite{liu2023pristi}, CSDI~\cite{tashiro2021csdi}, and DiffSTG~\cite{wen2023diffstg}, explicitly learn a data generation process based on diffusion or flow-based frameworks to model complex predictive distributions. Distribution parameterization methods, such as DER~\cite{amini2020deep} and SDER~\cite{wu2024evidence}, directly estimate the parameters of predictive distributions. To evaluate both point prediction accuracy and uncertainty estimation, we report MAE, RMSE, and MAPE, along with two standard probabilistic metrics: CRPS~\cite{matheson1976scoring} and MIS~\cite{wu2021quantifying}. Detailed descriptions of the metrics are provided in App.~\ref{Details of the baselines}.

\noindent \textbf{Implementation Details.} The proposed model and baselines are implemented using PyTorch 2.4.0 and trained on an NVIDIA GeForce RTX A4000 GPU. For each dataset, baseline hyperparameters are carefully tuned around recommended and default values to ensure optimal performance. All datasets are chronologically split into training, validation, and test sets with a 6:2:2 ratio, and Z-score normalization is applied using statistics from the training set. The model predicts the next 12 time steps of traffic data based on the previous 12 steps. We employ the Adam optimizer for model training, setting the batch size to 12 and the initial learning rate to 1e-4. For the PEMS03 dataset, the historical maximum traffic flow is used to approximate the road capacity, while for the other datasets, road capacity is calculated according to Eq.~\ref{Ca}. For the mean predictor, we select PDFormer~\cite{jiang2023pdformer}. Based on the hyperparameter study, we configure the number of principal components as K=3 and set the number of ST-Graph Blocks to 16. For the temporal attention, the number of heads is 12. The hidden representation dimensions are set to 32 for the PEMS03, PEMS04 and PEMS08 datasets, while for the Seattle dataset, the hidden representation dimension is increased to 64. We fix $\lambda_1 = 1$, and linearly increase $\lambda_2$ from 0 to 1 over epochs 20 to 50. The code is available at~\url{https://github.com/LvHaochenBANG/RIPCN.git}.
\setlength{\tabcolsep}{2.5pt}
% \footnotesize
\begin{table*}[ht]
\small
\centering
\caption{Overall forecasting performance comparison. Bold and underlined fonts indicate the best and second-best results.}
\vspace{0mm}
\label{PEMS04, PEMS08, PEMS03, and Seattle comparison}

\arrayrulecolor{black}
\begin{tabular}{c|c|c c c|c c c c|c c|c c} 
\toprule
  & Metric   & DeepAR   & Latent ODE & MC Dropout & STGNF & PriSTI       & CSDI         & DiffSTG        & DER & SDER & \textbf{RIPCN}  \\ 
\midrule
\multirow{5}{*}{\rotatebox{90}{PEMS03}} 
          & MAE       & 25.02{\scriptsize$\pm$0.26} & 24.11{\scriptsize$\pm$1.14} & 20.54{\scriptsize$\pm$0.72} & 19.94{\scriptsize$\pm$0.97} & 28.33{\scriptsize$\pm$1.27} & 18.92{\scriptsize$\pm$1.84} & 18.47{\scriptsize$\pm$0.59} & \underline{15.66{\scriptsize$\pm$0.66}} & 15.77{\scriptsize$\pm$0.63} & \textbf{15.28{\scriptsize$\pm$0.43}} \\
          & RMSE      & 40.66{\scriptsize$\pm$3.72} & 39.96{\scriptsize$\pm$6.20} & 33.79{\scriptsize$\pm$2.68} & 29.24{\scriptsize$\pm$3.02} & 41.96{\scriptsize$\pm$3.76} & 30.28{\scriptsize$\pm$4.97} & 32.27{\scriptsize$\pm$2.31} & \underline{26.58{\scriptsize$\pm$2.60}} & 26.79{\scriptsize$\pm$2.42} & \textbf{26.05{\scriptsize$\pm$2.08}} \\
          & MAPE      & 26.73{\scriptsize$\pm$2.75} & 17.73{\scriptsize$\pm$2.46} & 23.98{\scriptsize$\pm$1.04} & 24.21{\scriptsize$\pm$0.52} & 38.88{\scriptsize$\pm$4.84} & \underline{17.45{\scriptsize$\pm$0.53}} & 28.03{\scriptsize$\pm$0.06} & 18.75{\scriptsize$\pm$0.51} & 18.51{\scriptsize$\pm$0.49} & \textbf{12.07{\scriptsize$\pm$0.35}} \\
          & CRPS      & 0.1449{\scriptsize$\pm$0.0014} & 0.1091{\scriptsize$\pm$0.0010} & 0.1000{\scriptsize$\pm$0.0006} & 0.1117{\scriptsize$\pm$0.0010} & 0.1302{\scriptsize$\pm$0.0004} & \underline{0.0869{\scriptsize$\pm$0.0062}} & 0.0997{\scriptsize$\pm$0.0016} & 0.1242{\scriptsize$\pm$0.0012} & 0.1235{\scriptsize$\pm$0.0010} & \textbf{0.0778{\scriptsize$\pm$0.0012}} \\
          & MIS       & 1047.07{\scriptsize$\pm$86.76} & 964.42{\scriptsize$\pm$90.80} & 335.79{\scriptsize$\pm$183.41} & 377.06{\scriptsize$\pm$42.01} & 471.23{\scriptsize$\pm$26.03} & 224.81{\scriptsize$\pm$20.67} & \textbf{215.97{\scriptsize$\pm$17.65}} & 352.53{\scriptsize$\pm$21.64} & 411.79{\scriptsize$\pm$26.71} & \underline{221.34{\scriptsize$\pm$11.97}} \\
\midrule
\multirow{5}{*}{\rotatebox{90}{PEMS04}} 
          & MAE       & 27.46{\scriptsize$\pm$0.51} & 28.10{\scriptsize$\pm$1.07} & 23.60{\scriptsize$\pm$0.85} & 23.51{\scriptsize$\pm$0.91} & 28.30{\scriptsize$\pm$1.13} & 24.87{\scriptsize$\pm$1.36} & 21.90{\scriptsize$\pm$0.77} & \underline{21.11{\scriptsize$\pm$0.70}} & 21.18{\scriptsize$\pm$0.74} & \textbf{19.03{\scriptsize$\pm$0.87}} \\
          & RMSE      & 43.07{\scriptsize$\pm$1.93} & 43.95{\scriptsize$\pm$2.49} & 36.60{\scriptsize$\pm$1.64} & 38.64{\scriptsize$\pm$1.66} & 40.95{\scriptsize$\pm$1.94} & 38.27{\scriptsize$\pm$2.23} & 33.99{\scriptsize$\pm$1.52} & \underline{32.90{\scriptsize$\pm$1.44}} & 32.99{\scriptsize$\pm$1.49} & \textbf{30.19{\scriptsize$\pm$1.32}} \\
          & MAPE      & 19.46{\scriptsize$\pm$1.66} & 19.70{\scriptsize$\pm$1.57} & 17.46{\scriptsize$\pm$1.02} & 20.20{\scriptsize$\pm$1.00} & 23.63{\scriptsize$\pm$2.20} & 16.73{\scriptsize$\pm$0.73} & 16.54{\scriptsize$\pm$0.26} & 13.67{\scriptsize$\pm$0.38} & \underline{13.54{\scriptsize$\pm$0.31}} & \textbf{11.06{\scriptsize$\pm$0.22}} \\
          & CRPS      & 0.1260{\scriptsize$\pm$0.0037} & 0.1290{\scriptsize$\pm$0.0032} & 0.0912{\scriptsize$\pm$0.0008} & 0.0955{\scriptsize$\pm$0.0021} & 0.1027{\scriptsize$\pm$0.0008} & 0.0890{\scriptsize$\pm$0.0025} & \underline{0.0779{\scriptsize$\pm$0.0004}} & 0.0962{\scriptsize$\pm$0.0009} & 0.1103{\scriptsize$\pm$0.0011} & \textbf{0.0701{\scriptsize$\pm$0.0010}} \\
          & MIS       & 336.74{\scriptsize$\pm$24.76} & 1124.23{\scriptsize$\pm$98.11} & 411.93{\scriptsize$\pm$31.49} & 319.03{\scriptsize$\pm$25.77} & 376.18{\scriptsize$\pm$20.94} & \underline{259.98{\scriptsize$\pm$20.73}} & 257.85{\scriptsize$\pm$21.02} & 320.74{\scriptsize$\pm$22.15} & 424.33{\scriptsize$\pm$25.61} & \textbf{219.98{\scriptsize$\pm$17.88}} \\
\midrule
\multirow{5}{*}{\rotatebox{90}{PEMS08}} 
          & MAE       & 27.03{\scriptsize$\pm$0.53} & 26.08{\scriptsize$\pm$0.65} & 17.83{\scriptsize$\pm$0.19} & 20.82{\scriptsize$\pm$0.48} & 20.37{\scriptsize$\pm$0.17} & 19.61{\scriptsize$\pm$0.17} & 17.74{\scriptsize$\pm$0.16} & 16.66{\scriptsize$\pm$0.15} & \underline{16.46{\scriptsize$\pm$0.13}} & \textbf{15.14{\scriptsize$\pm$0.12}} \\
          & RMSE      & 41.73{\scriptsize$\pm$0.34} & 39.60{\scriptsize$\pm$0.82} & 27.80{\scriptsize$\pm$0.39} & 31.43{\scriptsize$\pm$0.66} & 30.03{\scriptsize$\pm$0.09} & 30.47{\scriptsize$\pm$0.09} & 27.16{\scriptsize$\pm$0.20} & 25.79{\scriptsize$\pm$0.23} & \underline{25.51{\scriptsize$\pm$0.20}} & \textbf{23.76{\scriptsize$\pm$0.18}} \\
          & MAPE      & 16.84{\scriptsize$\pm$0.96} & 17.71{\scriptsize$\pm$0.86} & 13.28{\scriptsize$\pm$0.43} & 16.64{\scriptsize$\pm$0.77} & 13.27{\scriptsize$\pm$0.29} & 12.24{\scriptsize$\pm$0.14} & 11.11{\scriptsize$\pm$0.23} & 10.43{\scriptsize$\pm$0.19} & \underline{10.22{\scriptsize$\pm$0.17}} & \textbf{9.27{\scriptsize$\pm$0.15}} \\
          & CRPS      & 0.1171{\scriptsize$\pm$0.0047} & 0.1129{\scriptsize$\pm$0.0058} & 0.0647{\scriptsize$\pm$0.0022} & 0.0644{\scriptsize$\pm$0.0044} & 0.0692{\scriptsize$\pm$0.0020} & 0.0668{\scriptsize$\pm$0.0020} & \underline{0.0607{\scriptsize$\pm$0.0025}} & 0.0743{\scriptsize$\pm$0.0019} & 0.0810{\scriptsize$\pm$0.0021} & \textbf{0.0565{\scriptsize$\pm$0.0019}} \\
          & MIS       & 261.77{\scriptsize$\pm$18.97} & 364.42{\scriptsize$\pm$43.81} & 307.29{\scriptsize$\pm$33.56} & 221.74{\scriptsize$\pm$20.18} & 258.86{\scriptsize$\pm$21.28} & 219.80{\scriptsize$\pm$19.32} & \underline{215.96{\scriptsize$\pm$19.24}} & 339.28{\scriptsize$\pm$21.86} & 404.21{\scriptsize$\pm$23.79} & \textbf{206.91{\scriptsize$\pm$19.03}} \\
\midrule
\multirow{5}{*}{\rotatebox{90}{Seattle}} 
          & MAE       & 112.67{\scriptsize$\pm$16.68} & 193.59{\scriptsize$\pm$17.70} & 144.79{\scriptsize$\pm$16.85} & 103.96{\scriptsize$\pm$16.88} & 185.59{\scriptsize$\pm$7.52} & 158.21{\scriptsize$\pm$8.65} & 127.05{\scriptsize$\pm$17.14} & 103.94{\scriptsize$\pm$15.33} & \underline{101.73{\scriptsize$\pm$14.28}} & \textbf{94.97{\scriptsize$\pm$14.59}} \\
          & RMSE      & 189.43{\scriptsize$\pm$30.58} & 330.67{\scriptsize$\pm$18.64} & 224.65{\scriptsize$\pm$28.89} & 173.62{\scriptsize$\pm$19.70} & 248.15{\scriptsize$\pm$14.46} & 223.54{\scriptsize$\pm$16.24} & 203.83{\scriptsize$\pm$23.95} & 161.79{\scriptsize$\pm$21.84} & \underline{156.32{\scriptsize$\pm$20.67}} & \textbf{147.74{\scriptsize$\pm$23.02}} \\
          & MAPE      & \underline{24.21{\scriptsize$\pm$6.56}} & 49.04{\scriptsize$\pm$7.58} & 34.12{\scriptsize$\pm$5.88} & 30.83{\scriptsize$\pm$7.32} & 57.98{\scriptsize$\pm$4.98} & 38.77{\scriptsize$\pm$4.09} & 27.47{\scriptsize$\pm$5.87} & 27.79{\scriptsize$\pm$5.42} & 26.45{\scriptsize$\pm$4.88} & \textbf{18.98{\scriptsize$\pm$3.76}} \\
          & CRPS      & 0.1331{\scriptsize$\pm$0.0234} & 0.2287{\scriptsize$\pm$0.0276} & 0.1432{\scriptsize$\pm$0.0239} & 0.1103{\scriptsize$\pm$0.0248} & 0.1744{\scriptsize$\pm$0.0133} & 0.1407{\scriptsize$\pm$0.0133} & \underline{0.1048{\scriptsize$\pm$0.0200}} & 0.1202{\scriptsize$\pm$0.0181} & 0.1269{\scriptsize$\pm$0.0167} & \textbf{0.0955{\scriptsize$\pm$0.0051}} \\
          & MIS       & 4506.59{\scriptsize$\pm$187.33} & 4743.74{\scriptsize$\pm$201.42} & 3135.79{\scriptsize$\pm$200.83} & \underline{1491.31{\scriptsize$\pm$87.67}} & 2914.08{\scriptsize$\pm$98.67} & 1563.50{\scriptsize$\pm$82.90} & 1761.73{\scriptsize$\pm$90.67} & 1533.50{\scriptsize$\pm$88.24} & 1601.55{\scriptsize$\pm$91.44} & \textbf{1297.79{\scriptsize$\pm$77.25}} \\
\bottomrule
\end{tabular}
\vspace{0mm}
\end{table*}
\setlength{\tabcolsep}{6.5pt}
\begin{table}[h]
\small
\vspace{0mm}
\caption{Ablation study results on Seattle dataset.}
\vspace{0mm}
\label{ablation}
\centering
\begin{tabular}{c|c|c|c|c|c}
\toprule
Variant  & MAE   & RMSE  & MAPE & CRPS & MIS \\ 
\midrule
w/o Impedance & 97.05 & 150.61 & 19.21   & 0.0979 & 1328.53\\
w/o ST-Graph  & 104.94 & 159.38 & 21.32   & 0.1141 & 1954.54\\
w/o $\mathcal{L}_R$  & 96.47 & 148.65 & 19.89   & 0.0977 & 1302.30\\
w/o $\mathcal{L}_D$  & 110.70 & 165.97 & 21.85 & 0.1247 & 2278.86 \\
w/o $\mathcal{L}_{V}$       & 95.92 & 148.67 & 19.60   & 0.0969 & 1356.27\\
RIPCN          & \textbf{94.97} & \textbf{147.74} & \textbf{18.98} & \textbf{0.0955} & \textbf{1297.79}\\
\bottomrule
\end{tabular}
\vspace{0 mm}
\end{table}
\vspace{-0mm}
\subsection{Performance Comparison}
Tab.~\ref{PEMS04, PEMS08, PEMS03, and Seattle comparison} presents the performance comparison of RIPCN with various probabilistic forecasting models. RIPCN consistently achieves the best results across all metrics on PEMS04, PEMS08, and Seattle, and ranks among the top two on PEMS03. For DeepAR and Latent ODE, their modeled distributions can be viewed as low-rank approximations of the target distribution, which limits their ability. MC dropout introduces stochastic dropouts, which can cause abnormal variations in the network and lead to poor predictions. Generative models like STGNF, CSDI, PriSTI, and DiffSTG typically exhibit stronger uncertainty quantification but suffer from biased mean predictions due to randomness in their initialization and sampling processes. In contrast, DER and SDER demonstrate high mean prediction accuracy but produce suboptimal uncertainty estimates due to weak supervision uncertainty signals and insufficient capacity to model complex distributions. 

In particular, the Seattle dataset yields significantly higher MAE and RMSE across all models. This is because it aggregates traffic flow on an hourly basis, resulting in longer prediction intervals and higher mean and variance, which increase forecasting difficulty. Despite this, RIPCN maintains strong point and uncertainty estimation performance, highlighting its robustness. In such challenging conditions, its advantages become even more evident.

\subsection{Ablation Study}
\label{ablation study}
To illustrate the contribution of each component within RIPCN, we conduct a comparison between RIPCN and its four variants: 
\begin{itemize}[leftmargin=*]
    \item \textbf{w/o Impedance}: Removes the impedance evolution network, using only the actual road adjacency matrix.

    \item \textbf{w/o ST-Graph}: Replaces the ST-Graph block with a fully connected layer.

    \item \textbf{w/o $\mathcal{L}_R$}: Removes the impedance MSE loss term $\mathcal{L}_R$ from the overall loss in Eq.~\ref{overall loss}.

    \item \textbf{w/o $\mathcal{L}_D$}: Removes the variance magnitude loss term $\mathcal{L}_D$.

    \item \textbf{w/o $\mathcal{L}_{V}$}: Removes the directional constraint loss term $\mathcal{L}_{V}$.
\end{itemize}

Tab.~\ref{ablation} presents the ablation results on the Seattle dataset to evaluate the impact of each component in RIPCN. \textit{w/o Impedance} results in noticeable performance degradation across all metrics, highlighting the importance of incorporating traffic transfer patterns into the model. When the impedance supervision loss $\mathcal{L}_R$ is removed, the model retains the impedance input but lacks explicit guidance. This comparison demonstrates that the impedance module itself plays a vital role, while the supervision loss $\mathcal{L}_R$ further enhances its effectiveness. \textit{w/o ST-Graph} causes the most significant performance drop. This demonstrates the critical role of structural modeling in capturing spatiotemporal PCs in traffic data.

Excluding the loss $\mathcal{L}_D$, which constrains the direction to align with the PCs, significantly degrades the quality of uncertainty estimation. In addition, removing the variance loss $\mathcal{L}_V$, which penalizes variance along PCs, results in wider prediction intervals, as reflected in the increase in MIS. These results collectively show that both direction-level and variance-level constraints are essential for high-quality probabilistic forecasting. The full model, RIPCN, achieves the best performance across all metrics,  confirming the effectiveness of the overall design.

\subsection{Hyperparameter Study}
\begin{figure}[h]
    \vspace{-0mm}
  \centering  \includegraphics[width=0.95\linewidth]{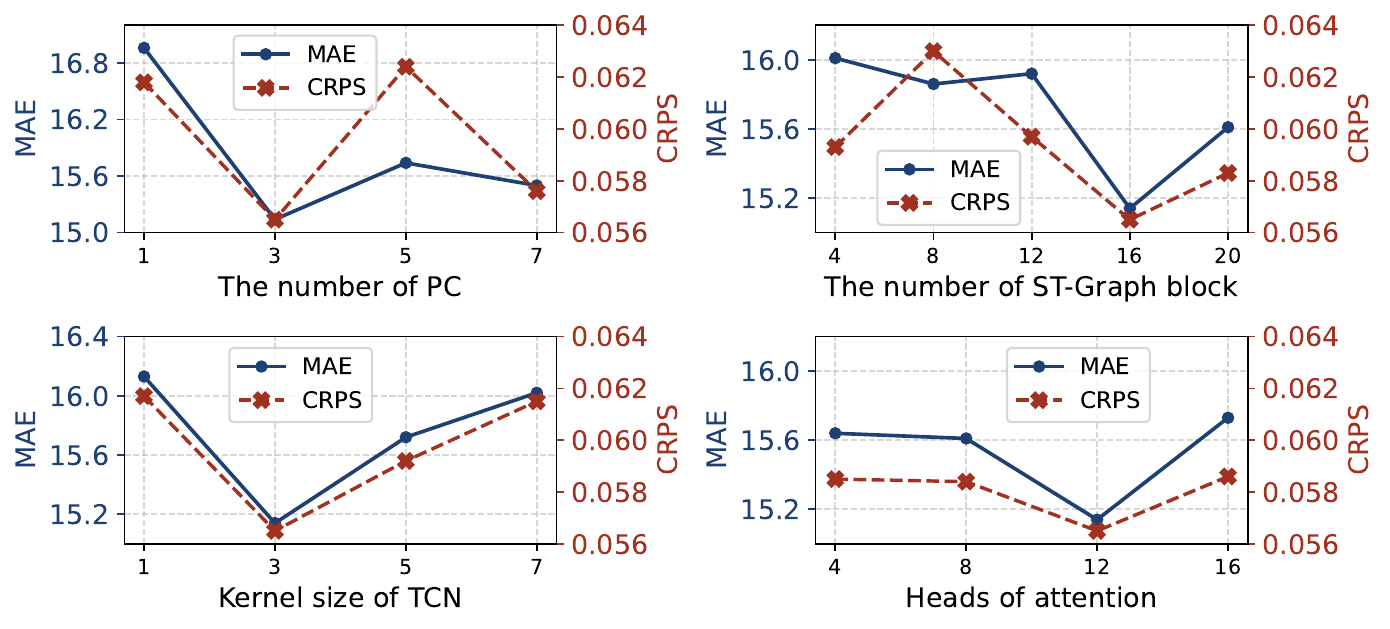}
  \vspace{-0mm}
  \caption{Influence of hyperparameters on PEMS08 dataset.}
  \Description{hyperparameter}
  \label{fig:hyperparam}
  \vspace{-0mm}
\end{figure}

We conduct hyperparameter experiments for RIPCN on the PEMS08 dataset, as shown in Fig.~\ref{fig:hyperparam}. The results show that choosing $K=3$ yields the best performance. RIPCN is generally robust to different values of $K$ because the first few PCs capture the dominant spatiotemporal uncertainty, with the first component contributing the most variance. Even $K=1$ provides reasonable uncertainty estimates, and additional components offer only limited gains. Using too many low-variance PCs introduces noise and may cause overfitting. Therefore, selecting $K\approx 3$ achieves a good balance between accuracy and efficiency. Additionally, we find that the optimal number of ST-Graph Blocks is 16. When both the historical and prediction time steps are set to 12, the TCN kernel size is best set to 3, and the optimal number of heads for the temporal attention is 12.

\subsection{Case Study}
\begin{figure}[h]
  \centering
\vspace{-0mm}
\includegraphics[width=\linewidth]{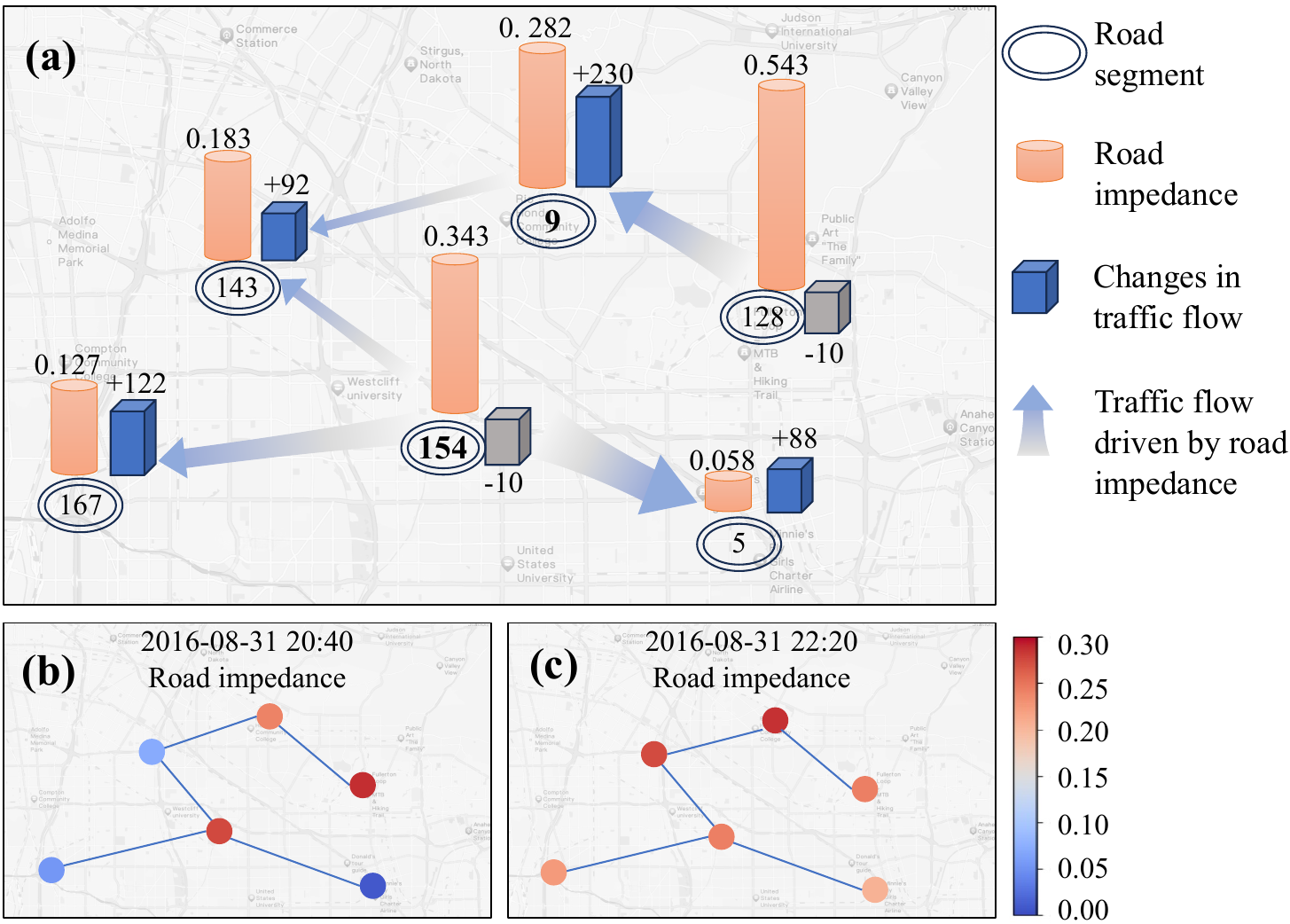}
  \vspace{-0mm}
  \caption{(a) shows the road impedance for PEMS08 segments over 100 minutes. (b) and (c) display the variations in road impedance during this period.}
  \Description{Traffic flow and corresponding road impedance.}
  \vspace{-0mm}
  \label{flow-impedance}
\end{figure}

\begin{figure}[h]
  \centering
\vspace{-0mm}
\includegraphics[width=0.90\linewidth]{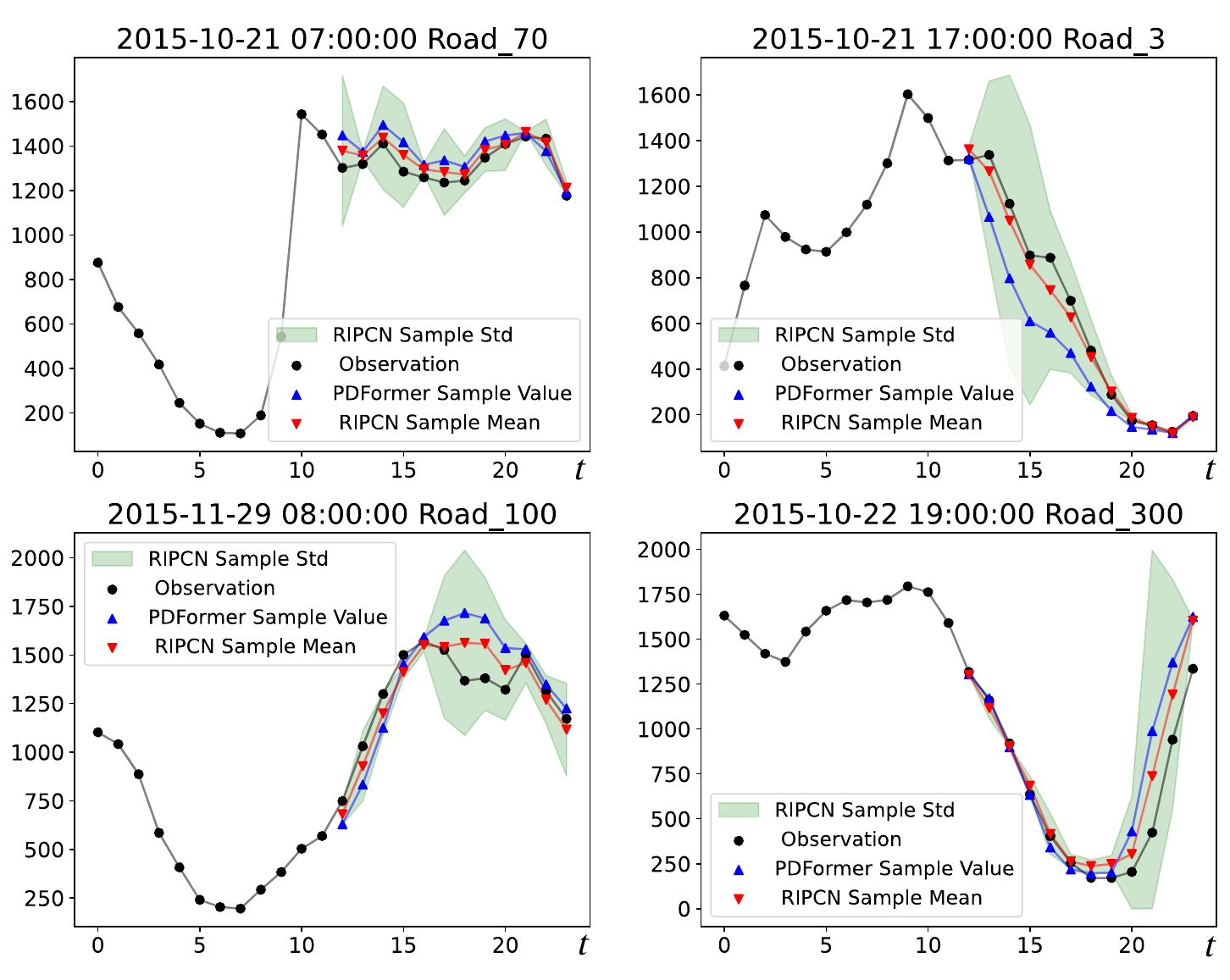}
  \vspace{-0mm}
  \caption{Traffic flow forecasting curve on Seattle dataset.}
  \Description{Four curves.}
  \label{curve}
  \vspace{-0mm}
\end{figure}

\begin{figure}[h]
  \centering
\vspace{-0mm}
\includegraphics[width=0.95\linewidth]{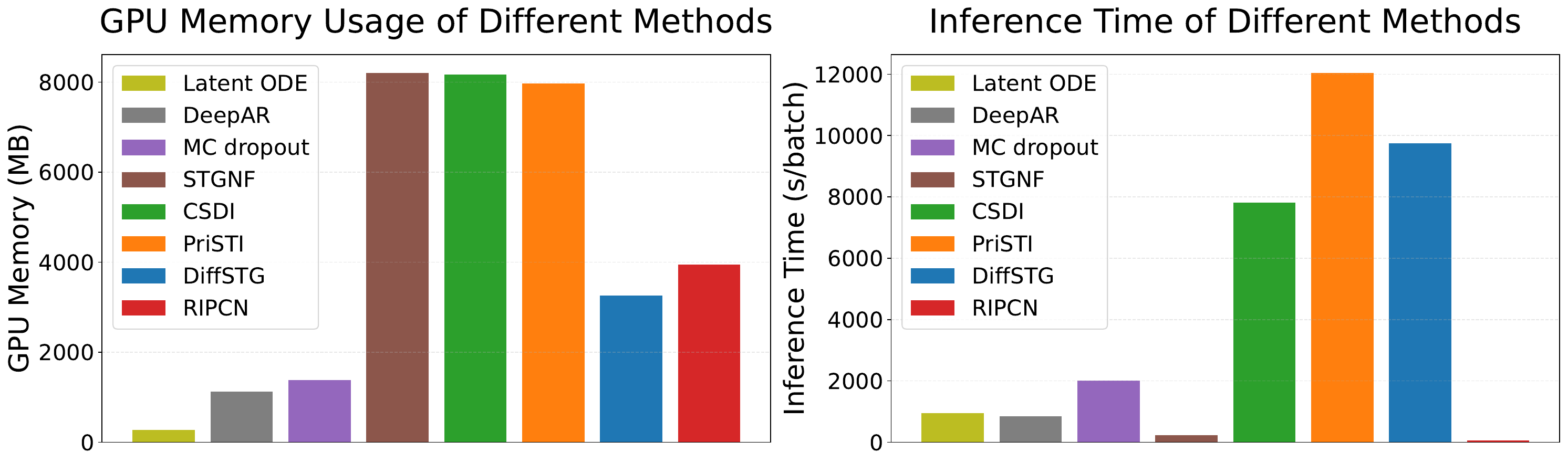}
  \vspace{-0mm}
  \caption{GPU memory and inference time comparison.}
  \label{efficiency}
  \vspace{-0mm}
\end{figure}

\label{case}
In this section, we present a case study to demonstrate the interpretability of our proposed RIPCN model. Fig.\ref{flow-impedance}(a) illustrates how impedance differences drive traffic flow transfer. For instance, based on the predicted impedance patterns, segment 154 is expected to exhibit outflow toward surrounding roads, and indeed shows a decrease in actual flow over time. In contrast, segment 9 experiences stronger inflow forces, resulting in an increase in flow. This demonstrates how road impedance influences flow dynamics. Fig.\ref{flow-impedance}(b)(c) further depict temporal variations in impedance, revealing a trend of flow shifting from high-impedance roads to nearby low-impedance ones.

Fig.~\ref{curve} shows the forecasting curves during peak periods and sudden changes, where deterministic models often deviate markedly from the ground truth. RIPCN still provides more accurate point estimates and uncertainty predictions that cover the true values, demonstrating its robustness and reliability. In subplot 3, RIPCN predicts low uncertainty during the steady rise before time step 15, and higher uncertainty following a sharp drop, with the predictive interval covering the true values. This highlights RIPCN’s ability to accurately capture elevated uncertainty while avoiding overestimation under stable conditions, enabled by its principal component module. App.~\ref{compare_var} compares RIPCN’s compact predictive intervals with the inflated variances of other models.

\subsection{Efficiency Study}
\setlength{\tabcolsep}{3.5 pt}
\begin{table}[h]
\centering
\caption{Efficiency comparison between RIPCN and the full spatio-temporal covariance model.}
\label{tab:fullcov}
\begin{tabular}{c|c|c|c|c}
\toprule
Dataset & Method & GPU (MB) & Train Time (s) & Infer Time (s) \\
\midrule
PEMS08  & Full Cov & 8346 & 287.85 & 229.94 \\
PEMS08  & RIPCN    & 3952 & 117.84 & 16.77  \\
Seattle & Full Cov & 19380 & 356.48 & 316.23 \\
Seattle & RIPCN    & 8364  & 125.64 & 52.87  \\
\bottomrule
\end{tabular}
\end{table}

We compare RIPCN with a variant that directly supervises the full spatiotemporal covariance of future flow. As shown in Tab.~\ref{tab:fullcov}, RIPCN reduces GPU memory consumption by about 50\% and accelerates inference by nearly 80\% compared with the full covariance model. This improvement arises from replacing the full $\mathcal{O}((TN)^2)$ covariance storage and $\mathcal{O}((TN)^3)$ decomposition with a low-rank PCA representation of complexity $\mathcal{O}(KTN)$. Moreover, the full covariance target contains substantial noise and redundancy, resulting in noticeably worse predictive accuracy compared with RIPCN.

Fig.~\ref{efficiency} shows the efficiency of sampling-based probabilistic forecasting methods on the PEMS08 dataset. The inference time for RIPCN includes the time for the mean predictor to generate the mean prediction and the time for uncertainty estimation. The results show that STGNF, CSDI, and PriSTI have higher GPU memory usage. Diffusion-based models including CSDI, PriSTI, and DiffSTG have significantly longer inference times. This is because their sampling process requires multiple and time-consuming denoising steps. In contrast, RIPCN demonstrates a significant advantage in inference time while using low GPU memory, thanks to its ability to generate multiple principal components at once and efficiently construct multiple samples.

\section{Conclusion}
In this paper, we propose the Road Impedance Principal Component Network (RIPCN), a novel framework for probabilistic traffic flow forecasting. By integrating road impedance and principal components, RIPCN achieves interpretability, more accurate point and uncertainty estimates, and improved computational efficiency.

\section{Acknowledgments}
This work was supported by the Beijing Natural Science Foundation (Grant Nos. L252034 and 4242029).

\newpage
\bibliographystyle{ACM-Reference-Format}
\bibliography{ref}

\newpage
\appendix
\section{Theoretical Justification of Mean Correction via Spatiotemporal Principal Components}
\label{correction theory}

To understand how the spatiotemporal principal components contribute to improving the predicted mean, we provide a theoretical analysis. Specifically, we show that under ideal conditions, the learned principal directions can be used to correct the prior mean prediction toward the true value. This refinement process is driven by the interplay between the PCA-related loss terms and the sampling mechanism, enabling adaptive adjustment along dominant directions in the data.

\textbf{Directional alignment via $\mathcal{L}_D$ (defined in Eq.~\ref{lw}).}  
Under the unit norm constraint $\| \mathbf{w}_k \|^2_F = 1$, minimizing the loss
\[
\mathcal{L}_w = -\sum_{\mathbf{X} \in \mathcal{D}} \sum_{k=1}^{K} \left\langle \mathbf{w}_k, \mathbf{X} \right\rangle_F^2
\]
is equivalent to maximizing $\left\langle \mathbf{w}_k, \mathbf{X} \right\rangle_F^2$ for each principal component $\mathbf{w}_k$ and sample $\mathbf{X}$.

By the matrix Cauchy-Schwarz inequality:
\[
\left\langle \mathbf{w}_k, \mathbf{X} \right\rangle_F^2 \leq \|\mathbf{w}_k\|_F^2 \cdot \|\mathbf{X}\|_F^2 = \|\mathbf{X}\|_F^2,
\]
with equality if and only if $\mathbf{w}_k$ and $\mathbf{X}$ are aligned. Thus, the theoretical optimal solution is:
\[
\mathbf{w}_k^* = \pm \frac{\mathbf{X}}{\|\mathbf{X}\|_F}.
\]
We first consider the case $\mathbf{w}_k^* = \frac{\mathbf{X}}{\|\mathbf{X}\|_F}$, for which the inner product achieves its maximum:
\[
\left\langle \mathbf{w}_k^*, \mathbf{X} \right\rangle_F^2 = \left( \frac{\left\langle \mathbf{X}, \mathbf{X} \right\rangle_F}{\|\mathbf{X}\|_F} \right)^2 = \|\mathbf{X}\|_F^2.
\]

\textbf{Variance consistency via $\mathcal{L}_V$.}  
The variance loss $\mathcal{L}_V$ (defined in Eq.~\ref{lc}) encourages the predicted variance $\sigma_k^2$ to approximate the actual variance along the direction $\mathbf{w}_k$, given by:
\[
\sigma_k^2 = \left\langle \mathbf{w}_k, \mathbf{X} \right\rangle_F^2.
\]
Substituting the optimal direction $\mathbf{w}_k^* = \frac{\mathbf{X}}{\|\mathbf{X}\|_F}$ yields:
\[
\sigma_k^2 = \left\langle \frac{\mathbf{X}}{\|\mathbf{X}\|_F}, \mathbf{X} \right\rangle_F^2 = \|\mathbf{X}\|_F^2 \quad \Rightarrow \quad \sigma_k^* = \|\mathbf{X}\|_F.
\]

\textbf{Sampling refinement via predicted uncertainty.}  
According to the sampling equation (Eq.~\ref{sample-process}), the predicted target is:
\begin{align*}
\hat{\mathbf{X}}_p + t_k \sigma_k^* \mathbf{w}_k^*
&= \hat{\mathbf{X}}_p + t_k \|\mathbf{X}\|_F \cdot \frac{\mathbf{X}}{\|\mathbf{X}\|_F} \\
&= \hat{\mathbf{X}}_p + t_k \mathbf{X}.
\end{align*}
Given that $\mathbf{X} = \mathbf{X}_P - \hat{\mathbf{X}}_p$, we can further write:
\begin{align*}
\hat{\mathbf{X}}_p + t_k \mathbf{X}
&= \hat{\mathbf{X}}_p + t_k (\mathbf{X}_P - \hat{\mathbf{X}}_p) \\
&= (1 - t_k) \hat{\mathbf{X}}_p + t_k \mathbf{X}_P.
\end{align*}
Thus, when $t_k = 1$, the sampled prediction exactly recovers the ground truth:
\[
\hat{\mathbf{X}}_p + \mathbf{X} = \mathbf{X}_P.
\]

\textbf{Symmetric case for negative alignment.}  
Now consider the case $\mathbf{w}_k^* = -\frac{\mathbf{X}}{\|\mathbf{X}\|_F}$. The analysis is analogous, leading to:
\begin{align*}
\hat{\mathbf{X}}_p + t_k \sigma_k^* \mathbf{w}_k^*
&= \hat{\mathbf{X}}_p - t_k \mathbf{X},
\end{align*}
which recovers the target when $t_k = -1$:
\[
\hat{\mathbf{X}}_p + \mathbf{X} = \mathbf{X}_P.
\]

\textbf{Conclusion.}  
These derivations demonstrate that under ideal conditions, where the learned principal directions perfectly match the direction between the prior and true values, the sampled correction can exactly recover the ground truth. In practice, since $\mathbf{w}_k$ is only an approximation, the correction may not be perfect, but the prediction can still be adjusted toward $\mathbf{X}_P$. This confirms that principal components can effectively guide mean correction.

Moreover, the role of $t_k$ is crucial. As the principal directions obtained from $\mathcal{L}_D$ may be aligned or anti-aligned with the data, $t_k$ must adapt its sign accordingly. In addition, the magnitude of $t_k$ can be adjusted in practice to optimize the degree of adjustment and reflect predictive uncertainty.

\section{Details of the baselines and metrics}
\label{Details of the baselines}
For each dataset, baseline hyperparameters are tuned based on recommended and default values to ensure optimal performance. Since both DER and SDER focus on loss function design for uncertainty modeling, we unify the backbone architecture by employing PDFormer~\cite{jiang2023pdformer} in all settings to ensure a fair and consistent comparison.

As for metrics, we use the Continuous Ranked Probability Score (CRPS)~\cite{matheson1976scoring} and the Mean Interval Score (MIS)~\cite{wu2021quantifying}. 

MIS is defined for estimated upper and lower confidence bounds. For a one dimensional random variable $Z\sim\mathbb{P}_Z$, if the estimated upper and lower confidence bounds are $u$ and $l$, where $u$ and $l$ are the $(1-\frac{\rho}{2})$ and $\frac{\rho}{2}$ quantiles for the $(1 - \rho)$ confidence interval, MIS is defined using samples $z_i\sim\mathbb{P}_Z$:
\begin{align*}
\text{MIS}_N(u,l;\rho)=\frac{1}{N}\sum_{i = 1}^{N}\left\{(u - l)+\frac{2}{\rho}(z_i - u)\mathbb{I}\{z_i>u\}\right. &\\
\left.+\frac{2}{\rho}(l - z_i)\mathbb{I}\{z_i<l\}\right\}&.
\end{align*}
We use the MIS suggested in~\cite{wu2021quantifying} for 95\% confidence intervals. The upper and lower bounds are determined by the 97.5th percentile and the 2.5th percentile of the sampled data.

CRPS is a metric to evaluate the performance of probabilistic prediction, which is commonly used to measure the compatibility of an estimated probability distribution $F$ with an observation $x$:
\[
\text{CRPS}(F,x)=\int_{\mathbb{R}}(F(z)-\mathbb{I}\{x\leq z\})^2\mathrm{d}z,
\]
where $\mathbb{I}\{x\leq z\}$ is an indicator function which equals one if $x\leq z$, and zero otherwise. Smaller CRPS means better performance. We uniformly approximate CRPS using a set of fixed quantiles (ranging from 0.05 to 0.95 with a step size of 0.05), where the prediction error at each quantile is measured by the corresponding pinball loss (i.e., quantile loss) to evaluate the discrepancy between the predicted distribution and the ground truth. 

For DER and SDER, we construct predictive distributions from the output parameters and sample 50 instances from each to estimate MIS and CRPS. In contrast, other methods inherently produce multiple samples during inference, enabling direct computation of MIS and CRPS via quantile-based approximation.

\section{Visualization of the predicted principal components}
\begin{figure}[h]
  \centering
  \includegraphics[width=0.8\linewidth]{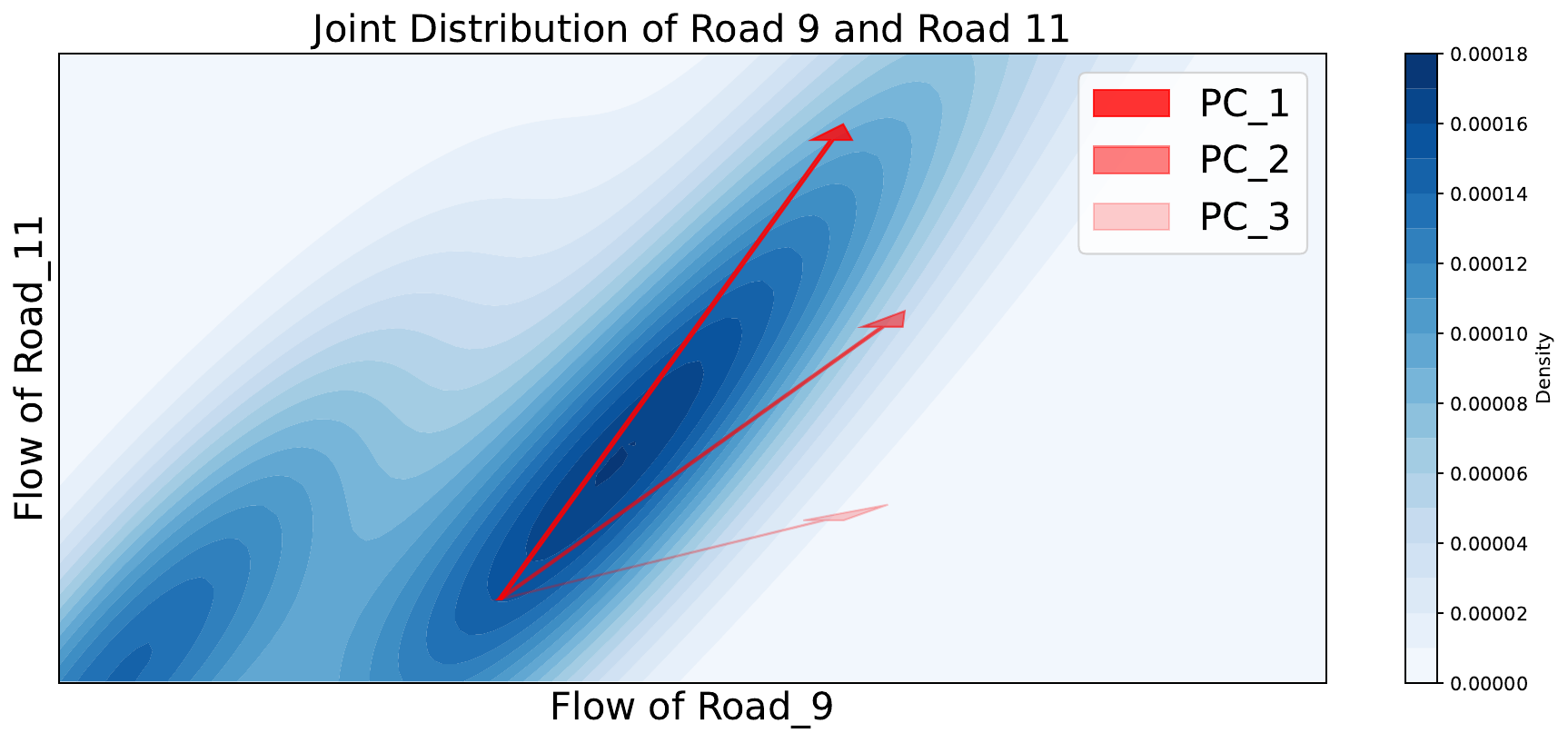}
  \vspace{-0mm}
  \caption{The probability distributions of Road 9 and 11 in Seattle and the PCs predicted by RIPCN.}
  \Description{predicted-PC}
  \label{predicted-PC}
  \vspace{-0mm}
\end{figure}

Fig.~\ref{predicted-PC} presents the real probability distributions of traffic flow for Roads 9 and 11 in the Seattle dataset alongside the principal components (PCs) predicted by RIPCN. Notably, PC\_1 effectively captures the main direction of data fluctuation, the dominant axis along which the probability distribution extends, while the other two PCs show only slight deviations. This indicates that the first principal component successfully represents the primary mode of variability in the data, which is crucial for accurate uncertainty modeling.

\begin{figure}[h]
  \centering
  \includegraphics[width=\linewidth]{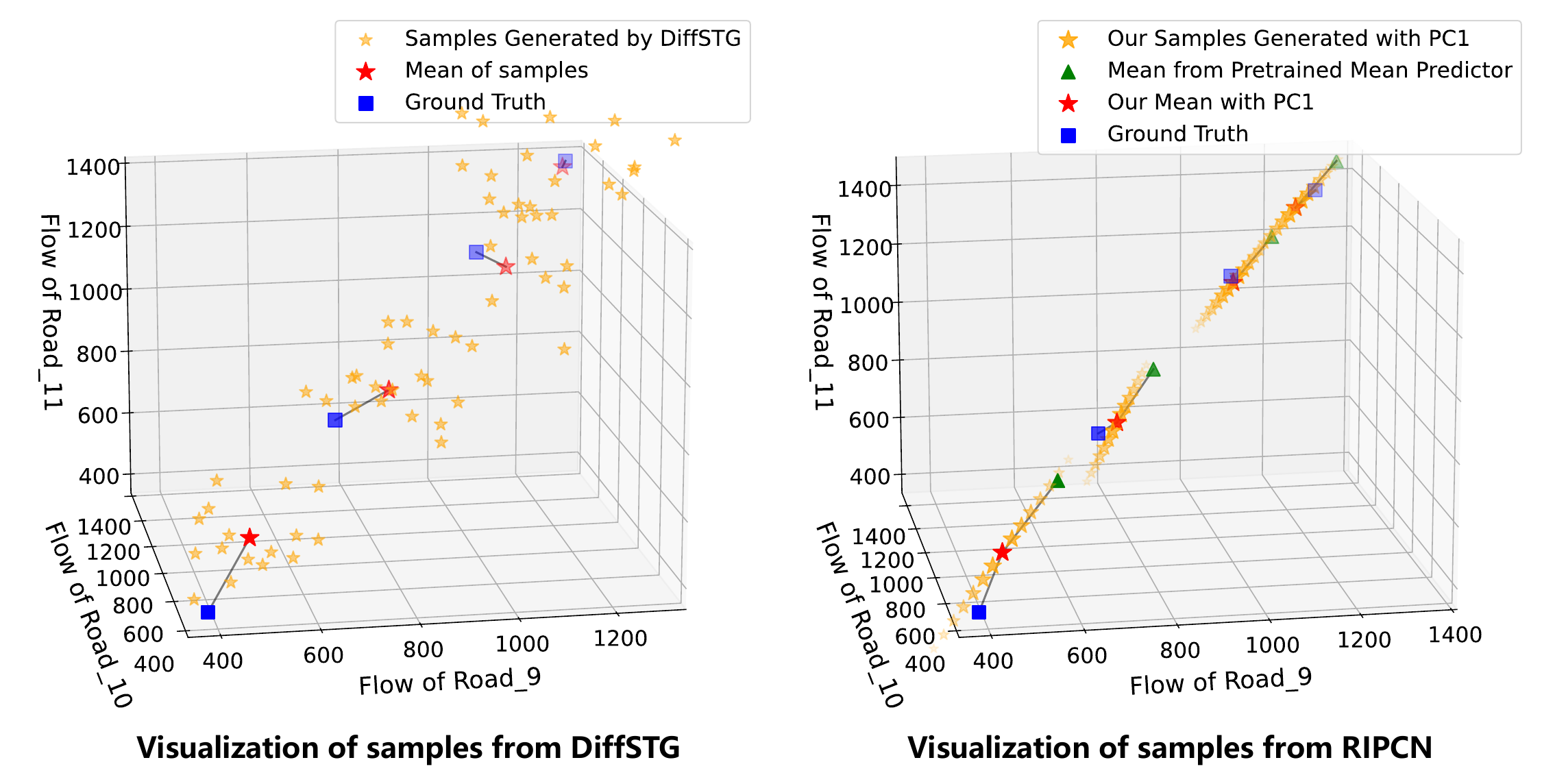}
  \vspace{-0mm}
  \caption{Visualization of samples from DiffSTG and RIPCN.}
  \Description{visual of diffstg and ripcn}
  \label{visual of diffstg and ripcn}
  \vspace{-0mm}
\end{figure}

The left part of Fig.~\ref{visual of diffstg and ripcn} visualizes the relationship among multiple sampled values, the predictive mean, and the true observed values for the DiffSTG model, depicted in a 3D plot for Roads 9, 10, and 11 across four consecutive time steps. Because DiffSTG does not explicitly consider the directionality of data fluctuations nor the spatiotemporal correlations inherent in uncertainty, its sampled values appear widely scattered without exhibiting meaningful correlation to the true values. This lack of structure results in an overestimation of predictive variance and less reliable uncertainty quantification.

In contrast, the right part of Fig.~\ref{visual of diffstg and ripcn} illustrates our RIPCN approach. It displays the mean predictions obtained from the pretrained mean predictor, the sampled values modulated by PC\_1, and the mean of these sampling predictions. Our method clearly captures the principal direction of joint fluctuations in the data with markedly higher accuracy. The sampled values are not randomly scattered but rather extend toward and encompass the true values, effectively aligning the uncertainty estimation with the actual joint variability observed between spatially and temporally correlated variables. This improved representation of uncertainty makes the model’s predictions more reliable.

Furthermore, by better capturing the direction of fluctuations, the mean of the sampled values is consistently closer to the true values compared to the original pretrained mean input to the model. This improvement reflects how modeling the principal components and their dynamics leads to refined point estimates as well as uncertainty quantification.

These advantages stem from RIPCN’s unique mechanism: it dynamically evolves the impedance graph to model realistic traffic flow transition patterns, and subsequently outputs principal components that encapsulate the spatiotemporal correlations of uncertainty. By integrating these domain-specific insights, RIPCN significantly enhances both the interpretability and accuracy of probabilistic traffic flow forecasting.

\section{The uncertainty of RIPCN vs. the uncertainty of overestimation}
\label{compare_var}
\begin{figure}[h]
  \centering
\includegraphics[width=\linewidth]{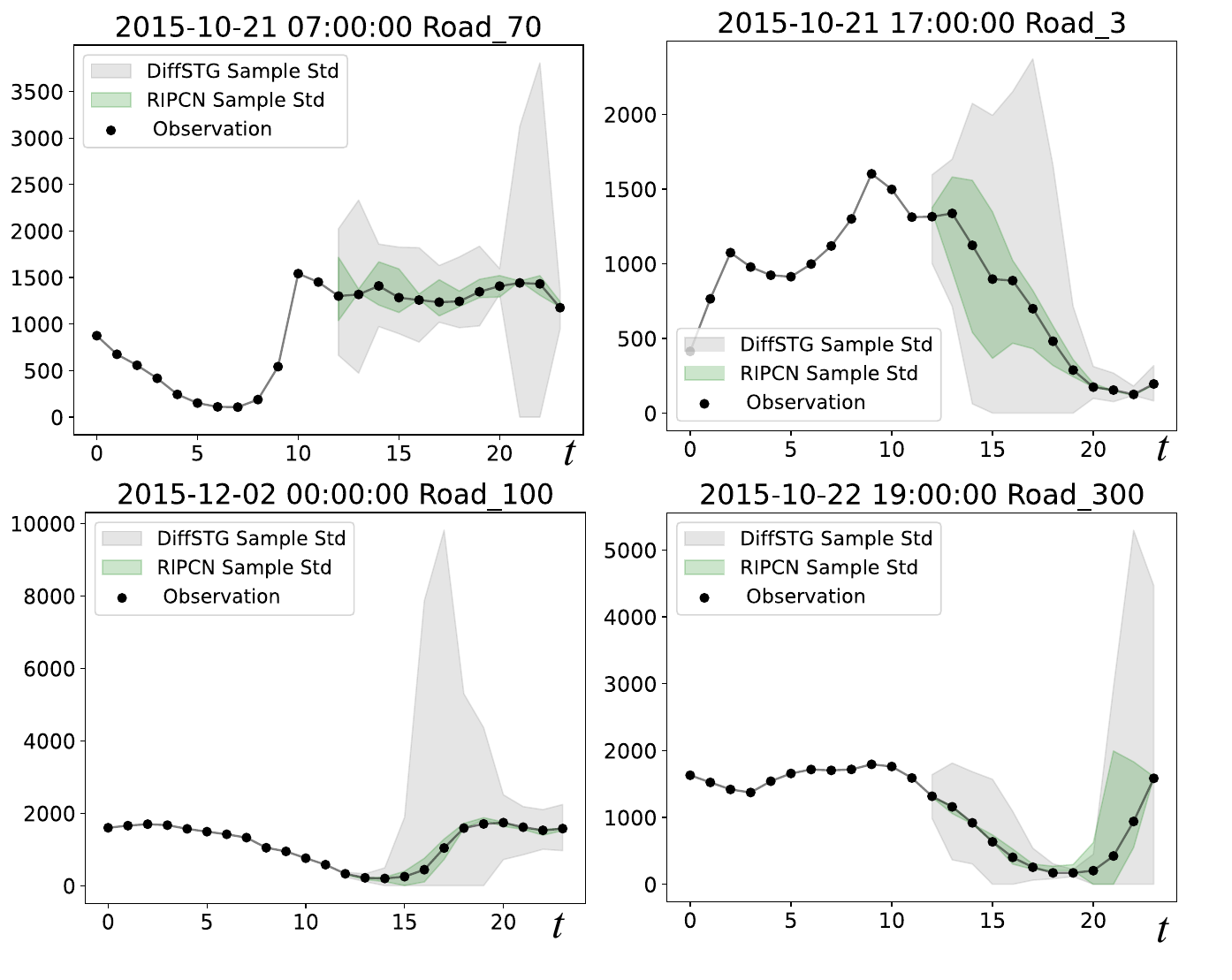}
  \vspace{-0mm}
  \caption{Comparison of the variances of DiffSTG and RIPCN}
  \Description{var-diffstg-ripcn}
  \label{var-diffstg-ripcn}
  \vspace{-0mm}
\end{figure}

Fig.~\ref{var-diffstg-ripcn} presents a comparison of the predictive variances produced by DiffSTG and RIPCN. DiffSTG exhibits consistently overestimated variances, while RIPCN produces compact and accurate variance estimates. Specifically, RIPCN assigns minimal variance during stable periods and appropriately larger variance during traffic surges and peak hours, effectively covering the true values. This behavior is of great practical significance, as it reflects the model’s ability to adapt its uncertainty based on the underlying traffic dynamics. In contrast, DiffSTG maintains high uncertainty estimates across all scenarios, regardless of actual variability. These observations highlight the importance of explicitly modeling the spatiotemporal correlation of uncertainty, which enables more precise and tighter uncertainty quantification.

\end{document}